\documentclass[10pt,twocolumn,letterpaper]{article}

\usepackage{iccv}
\usepackage{times}
\usepackage{epsfig}
\usepackage{graphicx}
\usepackage{amsmath}
\usepackage{amssymb}
\usepackage{pifont}
\newcommand{\cmark}{\ding{51}}%
\newcommand{\xmark}{\ding{55}}%
\usepackage{booktabs}
\usepackage{multirow}
\usepackage{dsfont}
\definecolor{iccvblue}{rgb}{0.21,0.49,0.74}
\usepackage[pagebackref,breaklinks,colorlinks,allcolors=iccvblue]{hyperref}

\usepackage[capitalize]{cleveref}
\crefname{section}{Sec.}{Secs.}
\Crefname{section}{Section}{Sections}
\Crefname{table}{Table}{Tables}
\crefname{table}{Tab.}{Tabs.}

\def\modelname{{GASP}}
\def\lidar{{lidar}}
\def\Lidar{{Lidar}}
\def\SOTA{{SotA}}
\def\aka{\emph{a.k.a}\onedot}
\def\pretrain{{pre-train}}
\def\Pretrain{{Pre-train}}
\def\pretrained{{{\pretrain}ed}}
\def\Pretrained{{{\Pretrain}ed}}
\def\pretraining{{{\pretrain}ing}}
\def\Pretraining{{{\Pretrain}ing}}

\def\finetuning{{fine-tuning}}

\newcommand{\parsection}[1]{\noindent\textbf{#1:}}
\usepackage{multirow}
\usepackage{textcomp}
\usepackage{tabulary}
\usepackage{colortbl}
\usepackage{siunitx}
\usepackage{graphicx}
\usepackage{nicematrix}
\definecolor{tabfirst}{rgb}{1, 0.7, 0.7} % red
\definecolor{tabsecond}{rgb}{1, 0.85, 0.7} % orange
\definecolor{tabthird}{rgb}{1, 1, 0.7} % yellow

\usepackage{gensymb}
\usepackage{mathtools}

\usepackage[symbol]{footmisc}

\definecolor{CornflowerBlue}{RGB}{100, 149, 237}
\definecolor{YellowGreen}{RGB}{154, 205, 50}
\definecolor{Apricot}{RGB}{251, 206, 177}

\def\fire{
{\raisebox{-2px}{\hspace{-2px}\includegraphics[height=1.1em, trim={0 0 0 0}, clip]{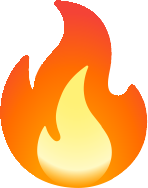}}\hspace{-2px}}
}

\def\snow{
{\raisebox{-2px}{\hspace{-2px}\includegraphics[height=1.1em, trim={0 0 0 0}, clip]{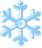}}\hspace{-2px}}
}

\title{GASP: Unifying Geometric and Semantic Self-Supervised \\ {\Pretraining} for Autonomous Driving}

\author{
    William Ljungbergh$^{*,1,2}$ \quad
    Adam Lilja$^{*,1,3}$ \quad
    Adam Tonderski$^{1,4}$ \quad
    Arvid Laveno Ling$^{1,3}$\\
    Carl Lindström$^{1,3}$\qquad
    Willem Verbeke$^{1}$\quad \hspace{5mm}
    Junsheng Fu$^{1}$\quad 
    Christoffer Petersson$^{1,3}$\\
    Lars Hammarstrand$^{3}$\quad
    Michael Felsberg$^{2}$ \vspace{2mm}\\
    \normalsize$^{1}$Zenseact \quad $^{2}$Linköping University \quad $^{3}$Chalmers University of Technology \quad $^{4}$Lund University\\
    {\tt\small \{firstname.lastname\}@\{zenseact.com, liu.se, chalmers.se\}}
}

\begin{document}
\maketitle
\footnotetext[1]{Denotes equal contribution.}
\begin{abstract}
Self-supervised {\pretraining} based on next-token prediction has enabled large language models to capture the underlying structure of text, and has led to unprecedented performance on a large array of tasks when applied at scale. 
Similarly, autonomous driving generates vast amounts of spatiotemporal data, alluding to the possibility of harnessing scale to learn the underlying geometric and semantic structure of the environment and its evolution over time. 
In this direction, we propose a \underline{g}eometric \underline{a}nd \underline{s}emantic self-supervised \underline{p}re-training method, \modelname, that learns a unified representation by predicting, at any queried future point in spacetime, (1) general occupancy, capturing the evolving structure of the 3D scene; (2) ego occupancy, modeling the ego vehicle path through the environment; and (3) distilled high-level features from a vision foundation model. 
By modeling geometric and semantic 4D occupancy fields instead of raw sensor measurements, the model learns a structured, generalizable representation of the environment and its evolution through time. 
We validate GASP on multiple autonomous driving benchmarks, demonstrating significant improvements in semantic occupancy forecasting, online mapping, and ego trajectory prediction. 
Our results demonstrate that continuous 4D geometric and semantic occupancy prediction provides a scalable and effective {\pretraining} paradigm for autonomous driving. For code and additional visualizations, see our \href{https://research.zenseact.com/publications/gasp/}{project page}.
\end{abstract}

\vspace{-2mm}
\section{Introduction}
%\vspace{-1mm}
Autonomous driving (AD) has the potential to improve safety, accessibility, and enhance transportation efficiency.
For an autonomous vehicle (AV) to operate safely and effectively, it must have a comprehensive understanding of its environment and the evolution thereof. In doing so, the AV must learn to reason about geometry and semantics in a dynamic environment.
\begin{figure}[t]
    \centering
    \includegraphics[width=\linewidth,keepaspectratio]{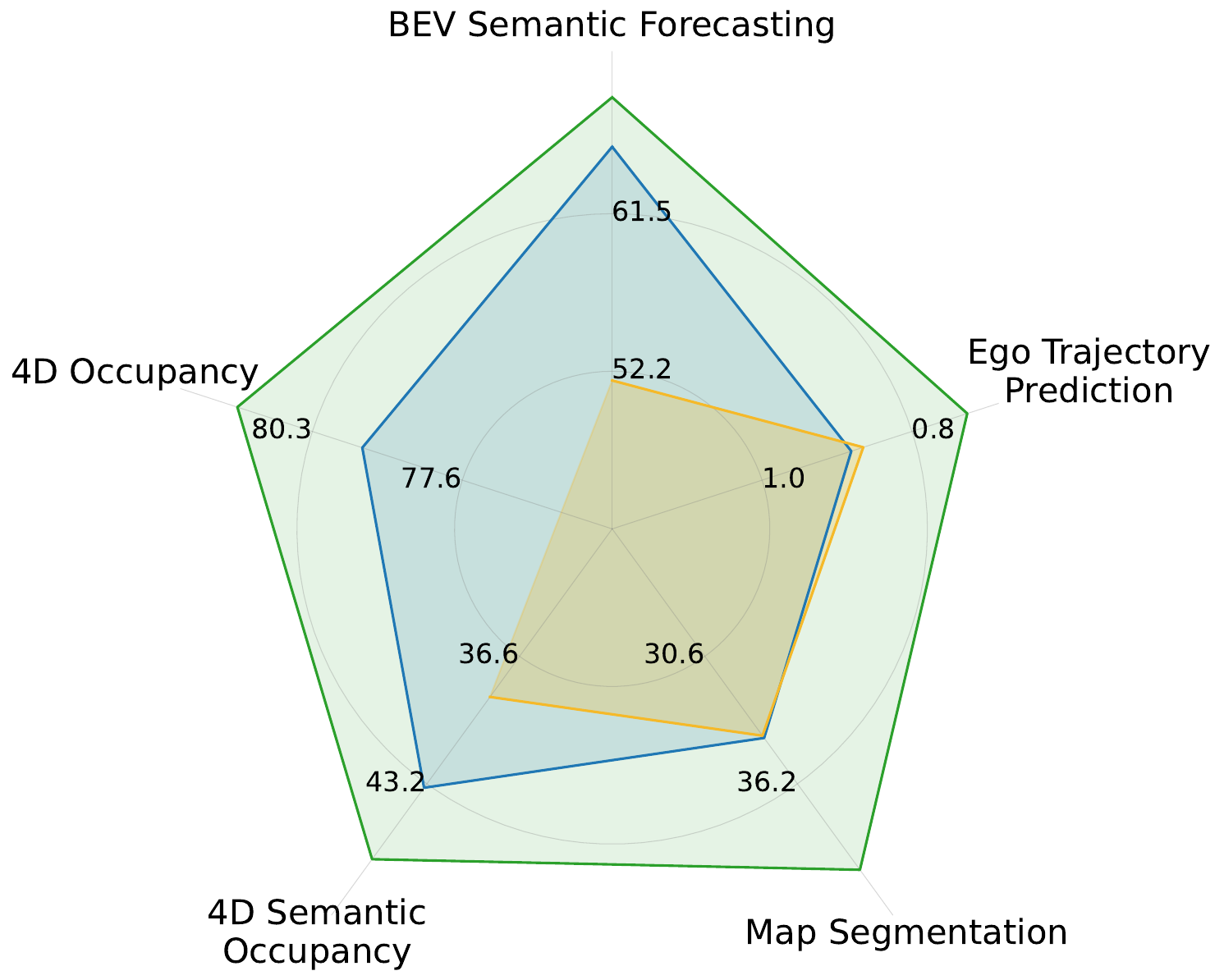}
    \caption{\colorbox{YellowGreen!50}{\modelname}~learns a structured, generalizable representation of the environment and its evolution and can be further trained to perform well on downstream AD tasks. 
    We outperform \SOTA~\pretraining~\colorbox{CornflowerBlue!50}{UnO}~\cite{agro2024uno} across the board, especially on primarily semantic tasks like map segmentation.
    \colorbox{Apricot!50}{No \pretraining} is displayed for reference. Downstream tasks requiring additional labels are post-trained using 1000 samples ($\sim$1\% of pre-training scale).}
    \label{fig:front-figure}
    \vspace{-6mm}
\end{figure}

To develop a comprehensive understanding of the environment, most existing systems rely heavily on large datasets with human-labeled annotations. 
Annotations are essential for solving tasks such as object detection and forecasting~\cite{yin2021center,jiang2023motiondiffuser}, online mapping~\cite{liao2024maptrv2}, and to enable multi-task frameworks with ego trajectory planning~\cite{hu2023_uniad, jiang2023vad, casas2021mp3}.
Unlabeled data is typically abundant when developing AD systems, but annotating a sufficiently diverse dataset is prohibitively expensive, limiting scalability of annotation reliant methods.
%To this end, self-supervised {\pretraining} has emerged as a critical enabler for scaling the performance of AV systems which typically have an abundance of raw, unlabeled, data available.

% Predictive learning over vast dataset, which has shown great success in language - chatgpt etc.

In other domains, \eg, natural language processing, self-supervised predictive learning over large datasets has been highly successful~\cite{radford2018improving, devlin-etal-2019-bert, brown2020language}. 
Researchers have explored predictive learning for AD, \eg by predicting future point clouds~\cite{Khurana_2023_CVPR, yang2024visual}, occupancy~\cite{zheng2024occworld, hu2023_uniad}, or video~\cite{Yang_2024_CVPR}.
These methods have shown promise but may struggle to model the continuous and dynamic nature of the driving environment, as they focus on predicting sensor observations rather than the underlying structure of the world.
Recent works~\cite{agro2023implicit, agro2024uno} address this by learning a representation in continuous spacetime. By predicting future occupancy from past \lidar~data, these methods offer a more accurate model of the inherently continuous real world.
%These methods are only supervised by future \lidar~data, which provides geometric and temporal cues but lacks the rich semantic understanding necessary for complex reasoning.
However, while future occupancy prediction provides strong geometric and temporal cues, it lacks the semantic richness needed comprehensive scene understanding and complex reasoning in downstream tasks.
%to capture the full complexity of sensor data for reasoning and downstream tasks.
% We propoose a method that integrates other sources of readliy available signals in order to improve the semantic representation in the learned model and show that this improves performance on downstream tasks.

To overcome this limitation, we propose {\modelname}, a self-supervised {\pretraining} method that integrates multiple sources of readily available signals in AV development: Future \lidar~scans, camera images, and ego poses.
By leveraging supervision from diverse sensor modalities, our method results in a richer representation of the environment and improves geometric, temporal, and semantic understanding.
Specifically, {\modelname} learns to predict occupancy, ego-path, and features from a vision foundation model~(VFM) in a continuous 4D (3D + time) representation.
The learned representation is useful on an array of downstream AD tasks, outperforming prior works as illustrated in \cref{fig:front-figure}.

% We also introduce several practical optimizations to improve efficiency, such as xyz.
Additionally, we introduce and demonstrate the efficacy of practical improvements:
 1) harvesting, negative information~\cite{KOCH200728}, from missing \lidar~rays for additional supervision, and
 2) a rotation augmentation strategy that significantly improves model generalization.
Our main contributions are:
\begin{itemize}
    \item Propose a self-supervised {\pretraining} method, \modelname, designed to learn a structured, generalizable 4D representation in continuous time by integrating geometric, temporal, and semantic supervision from multiple readily available signals.
    \item Demonstrate that \modelname {\pretraining} leads to improved generalization across multiple downstream autonomous driving tasks, significantly outperforming uni-modal {\pretraining} on tasks such as semantic occupancy forecasting, online mapping, and ego-trajectory prediction.
    \item Provide open-source code, including custom CUDA kernels for accelerated query generation and reimplementation of previously closed-source baselines, to facilitate further research in self-supervised learning for AD.
\end{itemize}

\section{Related work}
\begin{figure*}[t]
    \centering
    \includegraphics[width=\linewidth, trim={0 2mm 0 1mm}, clip]{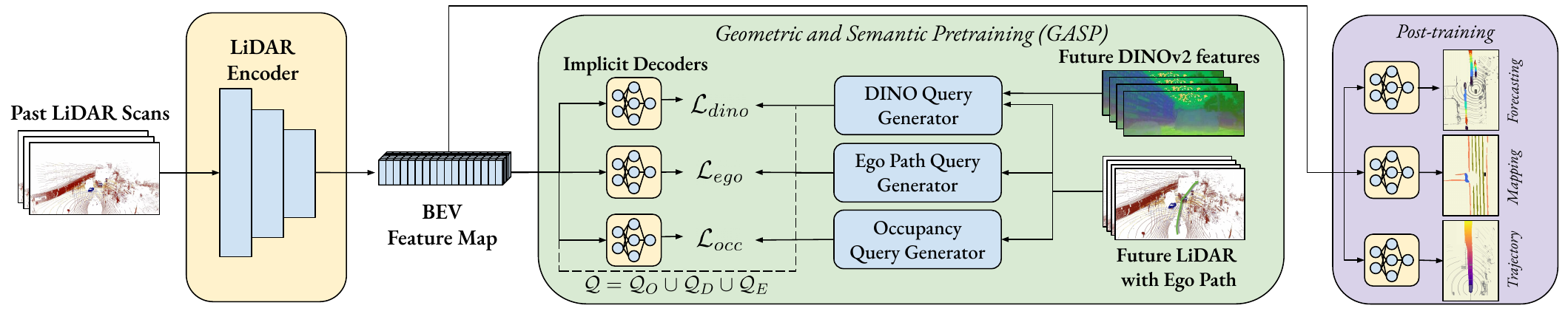}
    \caption{Overview of \modelname. Past \lidar\ scans are encoded into a BEV feature map. %using voxelization and a series of residual blocks. 
    These features are used by implicit decoders to predict DINOv2 features $\hat{\mathcal{D}}$, occupancy $\hat{\mathcal{O}}$, and ego-path  $\hat{\mathcal{E}}$ at the query points $\mathcal{Q}$ generated from future sensor data during \pretraining. 
    We also show that the learned representation is useful when transferred to an array of downstream AD tasks.}
    \label{fig:method-overview}
    %\vspace{-1mm}
\end{figure*}

Self-supervised learning has gained traction due to its ability to capture meaningful patterns without requiring expensive labels~\cite{dinov1, oquab2024dinov2, radford2021learning}, enabling greater scalability.
We apply these ideas to AD and provide an overview of the most relevant developments.

\parsection{Generative methods}
Generative methods withhold or alter parts of an input data sample and aim to reconstruct this part from the remaining data. Such methods learn features that generalize across a multitude of tasks. Masked input models have been applied to text~\cite{devlin-etal-2019-bert,NEURIPS2020_1457c0d6,radford2019language}, images~\cite{pathak2016context, he2022masked,bachmann2022multimae,bao2022beit,dosovitskiy2021an}, videos~\cite{Tong2022VideoMAEMA} and point clouds~\cite{hess2023masked,Yu_2022_CVPR, pang2022masked, liu2022masked}. These methods have been tailored to AD by jointly encoding multiple sensors and recovering masked inputs by neural rendering techniques~\cite{yang2024unipad, bevworld}.
Predicting future raw sensory data, such as point cloud forecasting~\cite{Khurana_2023_CVPR, yang2024visual} and video frame forecasting~\cite{Yang_2024_CVPR}, as a {\pretraining} step can also be seen in the light of masked input modeling. While such models learn relevant patterns in the data, they are also forced to learn details that are irrelevant for AD tasks: sensor intrinsics such as the scan pattern of a \lidar, and low-level stochastic information such as the lighting of each reconstructed pixel.
% Add n3ext-token prediction like Gaia?

\parsection{Implicit generative methods}
Alternatively, sensory data forecasting can be rephrased as generic occupancy forecasting \cite{agro2024uno}. 
This has two advantages compared to direct generative methods: Future occupancy depends on the dynamics of the environment but not on that of the sensors, and occupancy is directly useful for downstream tasks in AD. 
By encoding past sensory information (\eg, \lidar~\cite{agro2024uno, agro2023implicit,Khurana_2023_CVPR, zheng2024occworld} or images~\cite{yang2024visual, Ma_2024_CVPR}) into a latent representation they reason about the future at discrete~\cite{hu2023_uniad, zheng2024occworld, hu2021fiery, Khurana_2023_CVPR} or continuous~\cite{agro2024uno, agro2023implicit} times. 
We follow this trend, taking inspiration from~\cite{agro2024uno}, to implicitly predict a 4D \textit{continuous occupancy field} that can be queried at 4D coordinates $q = (x,y,z,t)$ to yield a local occupancy probability.
Our method predicts a continuous occupancy field, but extends this by implicitly predicting both the future path of the ego vehicle and the flow of a rich latent representation in a unified way.

\parsection{Embedded predictions}
By operating directly in the domain of abstract representations, unimportant and noisy low-level details can be ignored. 
Methods in this category often rely on contrastive learning~\cite{chen2020simple} or feature alignment between augmented views of the same input, as done in DINO~\cite{dinov1,oquab2024dinov2}.
An alternative is to use latent information to reconstruct missing parts of the input, which has shown promising results for images~\cite{ijepa} and videos~\cite{vjepa}. 
Building on these ideas, we encourage our model to implicitly predict high-level abstract features in the future, forcing it to reason about semantics and dynamics.
Rather than training a new image encoder, we distill features generated by DINOv2~\cite{oquab2024dinov2}, a model {\pretrained} on a large-scale dataset to produce generalizable image representations. 

\parsection{Trajectory planning}
Predicting a desirable future trajectory is the ultimate goal of an AV. 
Contemporary methods typically follow an end-to-end design, where intermediate outputs contribute to predicting a final drivable trajectory~\cite{casas2021mp3, liang2020pnpnet, hu2023_uniad, jiang2023vad, weng2024drive, tong2023scene}.
This structured approach improves ego trajectory forecasting and increases performance on intermediate tasks, but also relies on expensive labeled data~\cite{hu2023_uniad}. 
Trajectory prediction itself is a rich self-supervised signal that requires no human annotations. 
Therefore, we incorporate ego-path prediction as a {\pretraining} task to integrate end-to-end path prediction with future occupancy and semantic feature information, providing a richer understanding of driving scenes.

\parsection{Lifting vision foundation models to 3D}
Several works have explored lifting image features to 3D. Lifting CLIP features into 3D~\cite{vobecky2023pop, hess2024lidarclip} can enhance semantic understanding, while~\cite{ovsep2024better} combine CLIP and SAM~\cite{kirillov2023segment} for text-promptable point cloud segmentation. These approaches rely on full feature dimensionality, while~\cite{yang2023emernerf} demonstrate that a subset of DINOv2 features is sufficient to improve semantic understanding and enable few-shot auto-labeling in scene reconstruction.
With this insight, we distill positional embedding-denoised DINOv2 features~\cite{yang2024denoising}. A key distinction is that we predict these features' future evolution, capturing the representations' temporal dynamics.

\section{Method}
\label{sec:method}
We propose \modelname, a self-supervised method that trains a model to reason about the evolution of geometry and semantics in temporal data. The model is trained to predict future occupancy (geometry and time), vision foundation model (VFM) features (semantics and time), and ego-path (geometry and semantics) at any queried point in continuous spacetime. %within a region of interest $\mathcal{R}_I$. 
We outline the model architecture in~\cref{sec:model-architecture} and~\cref{fig:method-overview}, explain the {\pretraining} procedure in~\cref{sec:pretraining-procedure}, and how to enhance the model's usability by leveraging labeled data with post-training in~\cref{sec:post-training-procedure}.

\subsection{Model architecture}
\label{sec:model-architecture}
We adopt the model architecture in~\cite{agro2024uno,agro2023implicit}.
The model uses a \lidar~encoder to parametrize a feature field conditioned on past sensor data that can be queried for occupancy through a lightweight implicit decoder. 
In addition to that, we add additional decoders to predict VFM features, and ego-vehicle occupancy at any 4D point, see \cref{fig:method-overview}. 
We follow~\cite{agro2024uno} and use temporal \lidar\ data as input in this work, but note that the decoding architecture is sensor-agnostic.

The \lidar~encoder processes $K_{past}$ past \lidar~scans into a bird's-eye-view (BEV) feature map $Z \in \mathbb{R}^{H \times W \times C}$. 
Scans are aggregated with ego-motion compensation and voxelized~\cite{yang2018pixor} before being encoded by a ResNet-style~\cite{he2016deep} backbone with deformable attention~\cite{zhu2020deformable} and a Feature Pyramid Network~\cite{lin2017feature}.
The decoders query the BEV feature map $Z$ to predict target values through a lightweight architecture based on deformable attention~\cite{zhu2020deformable}, residual blocks, and a final linear layer. This design enables efficient parallel query decoding, while doing the heavy lifting in the encoder. 
We use the same architecture for all decoders heads to, for each query point $\mathbf{q}_i$ predict occupancy $\hat{o}_i = H_o(\mathbf{q}_i)$, VFM feature $\hat{v}_i=H_v(\mathbf{q}_i)$, and ego path query $\hat{e}_i=H_e(\mathbf{q}_i)$. 

\subsection{{\Pretraining} procedure}
% =====================
% UNDER RECONSTRUCTION
% =====================
\label{sec:pretraining-procedure}
For our self-supervised {\pretraining}, we generalize the approach of~\cite{agro2024uno} and produce a set of 4D (3D + time) $N$ data samples $\mathcal{D} = \{\langle\mathbf{q}_i, a_i \rangle\}_{i=0}^N$ comprising of queries $\mathbf{q}_i$ and targets $a_i$ from future data at $t \in [0, T_{max}]$. 
We assume temporal sequences of \lidar~data with known ego-vehicle motion throughout the sequence, standard in AD datasets~\cite{caesar2020nuscenes, nuplan, Argoverse2, Sun_2020_CVPR, alibeigi2023zenseact, xiao2021pandaset}.
We denote the set of $M$ \lidar~points with their corresponding sensor origin $\mathcal{P} = \{\langle \mathbf{p_i}, \mathbf{s}_i\rangle\}_{i=1}^M$, where each \lidar~point ${\mathbf{p_i}} = (x_i,y_i,z_i)$ and $\mathbf{s}_i = (x_i, y_i, z_i)$ has a corresponding time $t_i$ at which the ray was emitted.
We extend the geometric occupancy supervision, using data samples $\mathcal{D}_O$, with vision foundation model feature supervision from $\mathcal{D}_F$ and future ego path traversal probabilities using $\mathcal{D}_E$. 
We elaborate on the training procedure below.

\parsection{Occupancy data generation} 
We follow the methodology of \cite{agro2024uno} to create training samples for future occupancy prediction. \emph{Unoccupied} query points are sampled along the \lidar~ray up to the \lidar~return:
\begin{equation}
    \label{eq:negative-queries}
    \mathcal{D}_O^- = \{\langle \mathbf{s}_i + r (\mathbf{p}_i - \mathbf{s}_i) , 0\rangle \ | \  r \in (0, 1)  \}_{i=0}^N
\end{equation}
Positive, \emph{occupied}, queries are generated within a buffer zone with length $\delta$ behind the \lidar~return
\begin{equation}
    \label{eq:positive-queries}
     \mathcal{D}_O^+ =  \{ \langle \mathbf{p}_i + \frac{r(\mathbf{p}_i - \mathbf{s}_i)}{||\mathbf{p}_i - \mathbf{s}_i||} , 1 \rangle \ | \  r \in (0, \delta) \}_{i=0}^N
\end{equation}
In practice, we randomly select $N_O^+$ and $N_O^-$ from $\mathcal{D}_O^+$ and $\mathcal{D}_O^-$ respectively to form the data samples $\mathcal{D}_O$ to supervise future occupancy. 

\parsection{Vision foundation model data generation} 
To generate training samples $\mathcal{D}_F$ for learning temporal semantic features, we project future \lidar~points to the images closest in time, while compensating for ego-motion, and fetch the corresponding feature.
Since the \lidar~is typically mounted higher than the camera, its rays can pass over objects -- such as vehicles -- and hit the ground or other surfaces behind them.
Naively projecting onto the image, these may be assigned incorrect semantic features, leading to noisy supervision.
We therefore apply per-pixel min-depth filtering, ensuring that only the closest visible points, $\mathcal{P}_{\text{vis}} \subseteq \mathcal{P}$, contribute to training.
At the projected locations, we extract the feature $\mathbf{F}_i$ from the output of a frozen vision foundation model as the semantic training target:

\begin{equation}
    \label{eq:dino-queries}
     \mathcal{D}_F =  \{ \langle \mathbf{p}_i + \frac{r(\mathbf{p}_i - \mathbf{s}_i)}{||\mathbf{p}_i - \mathbf{s}_i||} , \mathbf{F}_i \rangle \ | \ r \in (0, \delta), \langle \mathbf{p_i}, \mathbf{s}_i\rangle \in \mathcal{P}_{\text{vis}}\}%_{i=0}^{|\mathcal{P}_{\text{vis}}|}
\end{equation}
In this work, we chose to use the denoising DINOv2 model~\cite{yang2024denoising} to mitigate known issues in lifting DINOv2 features with positional encodings~\cite{yang2023emernerf}. However, we note that features from any vision foundation model could be used.
The proposed procedure lifts information present in DINOv2 features from 2D to 3D, allowing for joint spatial and semantic reasoning.

\parsection{Ego path data generation} 
We generate ego path training samples from the future poses of the ego vehicle $\mathcal{E} = \{\mathbf{e}_i = (x_i,y_i,z_i)\}_{i=1}^{M^e}$, from which we define
the set of positive queries $\mathcal{Q}^+_E = \{\mathbf{q} \ | \ ||\mathbf{q}-\mathbf{e}_i|| \leq  w_{ego}\}_{i=1}^{M^e}$, 
as points closer than, $w_{ego}$, to the vehicle. This gives us the positive data samples 
\begin{equation}
\mathcal{D}_E^+ = \{\langle \mathbf{q} , 1 \rangle 
 | \ \mathbf{q}\in \mathcal{Q}^+_E\}\end{equation}
Negative samples are instead located in the rest of the space within the region of interest $\mathcal{R}_I$:
\begin{equation}
\mathcal{D}_E^- = \{\langle \mathbf{q} , 0 \rangle | \ \mathbf{q} \in \mathcal{R}_I \setminus \mathcal{Q}^+_E\}
\end{equation}

We emphasize the distinction between ego-path and ego-trajectory. 
The former has no notion of time, only positions. 
Focusing solely on the driven path avoids ambiguity that occurs when the ego-vehicle is stationary. 
The full positive sampling volume could technically be a valid path for the ego-vehicle to traverse. 
However, directly predicting it forces the model to learn an explicit multi-modal distribution of possible ego-paths. 
%Our formulation allows the task to be solved with a unified representation and decoder architecture together with the evolution of DINOv2 features and occupancy.
Our formulation allows the task to be solved within our unified framework, alongside the prediction of evolving occupancy and semantic features.

\begin{figure*}
    \centering
    \includegraphics[width=\linewidth]{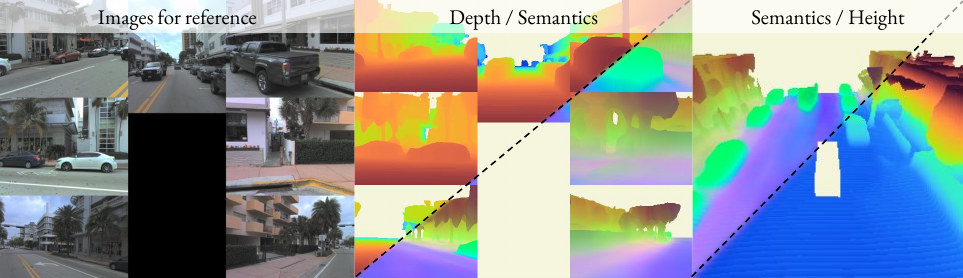}
    \caption{Predicted occupancy (colored by depth and height respectively) and DINOv2 features (mapped to RGB using the three most important features) projected into camera views, as well as a holistic view from slightly above and behind the ego vehicle. Different type of objects such as road, vehicles, buildings, and trees have different features, indicating the model has semantic understanding of the objects in the scene.
    The injected white box represents the ego vehicle for clarity.}
    \label{fig:dino-qual}
    %\vspace{-3mm}
\end{figure*}

\parsection{Training loss}
We train our model using a multi-task loss that consists of binary cross-entropy terms for occupancy 
$\mathcal{L}_{\text{occ}}$ and ego-path probabilities $\mathcal{L}_{\text{ego}}$, and $L1$-loss for DINOv2 features $\mathcal{L}_{\text{dino}}$.
The total loss is defined as:
\begin{equation}
    \mathcal{L} = \lambda_{\text{occ}}\mathcal{L}_{\text{occ}} + \lambda_{\text{dino}}\mathcal{L}_{\text{dino}} + \lambda_{\text{ego}}\mathcal{L}_{\text{ego}},
\end{equation}
where $\lambda_{\text{occ}}$, $\lambda_{\text{dino}}$, and $\lambda_{\text{ego}}$ are hyperparameters.

\parsection{Rotation augmentation}\label{sec:rot-aug}
%Most collected data comes from nominal driving scenarios, in which the ego vehicle and other road participants move in parallel to the road, \ie axis-aligned with the BEV representation of our model.
Real-world driving is inherently dominated by straight-road driving, where the motion of most road participants is axis-aligned with the ego coordinate system.
This has been shown to induce a strong bias in \eg online mapping~\cite{lindstrom2024nerfs}. 
We observed similar tendencies in the initial training of \modelname~and address this by randomly rotating the coordinate system by $\theta \in [\theta_{min}, \theta_{max}]$ during training. This reduces the directional bias and promotes a more diverse representation of motion.

\parsection{Missing \lidar~ray inference}\label{sec:missing-rays}
A \lidar~is an active sensor that measures distances by emitting laser rays. Unobstructed rays do not return measurements (\aka\ missing). Disregarded in most applications and datasets~\cite{caesar2020nuscenes, nuplan, Argoverse2, Sun_2020_CVPR, alibeigi2023zenseact, xiao2021pandaset}, missing rays carry valuable information about unoccupied space. Following~\cite{tonderski2024neurad}, where the utility of missing rays for learning scene geometry was demonstrated, we infer missing rays from \lidar~scans and leverage them to sample negative occupancy queries. Recovering individual missing rays is prone to false positives. To increase robustness, we adapt the algorithm to focus on identifying extended regions of missing rays.

\subsection{Post-training procedure}
\label{sec:post-training-procedure}
\modelname~aims to equip the model with a strong understanding of geometry, semantics, and dynamics. To assess the quality of the learned representations, we adapt the model or introduce additional task-specific heads during post-training (see~\cref{fig:method-overview}).
The learned representation $Z$ can be used in multiple ways:  querying it similarly to the {\pretraining} phase, or using $Z$ directly (or resampled) as input to another network.
This flexibility enables the straightforward addition of task-specific heads for a variety of downstream applications.

% instead explained in each experiment subsection
% We evaluate our learned representation on three downstream tasks: BEV occupancy forecasting, online mapping, and ego trajectory prediction.
% In BEV occupancy forecasting we finetune the query-based occupancy decoder to predict class-wise 2D occupancy, assigning a learned height to all queries.
% For online map segmentation, we use a lightweight U-Net-inspired decoder~\cite{harley2023simple}, which predicts lane dividers, road edges, and pedestrian crossings from $Z$.
% Ego trajectory forecasting uses a decoder architecture inspired by~\cite{zhu2020deformable}, which queries the BEV representation with predefined anchor trajectories. These generate multiple proposals for the ego vehicle’s future positions.
% !!POSSIBLY ADD 4D SEMANTIC OCCUPANCY HERE!!

\section{Experiments}
\label{sec:experiments}
In this section, we evaluate the proposed self-supervised objective and assess whether the model learns a generalizable representation of the environment and its evolution.

First, we evaluate the performance of the {\pretrained} model on \emph{Geometric 4D Occupancy Forecasting}~(\cref{sec:experiment-4d-occ}).
The {\pretrained} model's generalization capabilities are evaluated on downstream AD tasks: \emph{Semantic BEV Forecasting}~(\cref{sec:experiment-semfor}), \emph{Map Segmentation}~(\cref{sec:experiment-map-seg}), and \emph{Ego Trajectory Forecasting}~(\cref{sec:experiment-ego-trajectory}).
We study two settings:
1) Feature evaluation ({\snow}); We freeze the learned encoder and train only the head. 
This allows us to measure how relevant the information encoded in the BEV features is.
2) Full network adaptation (\fire); We train both the encoder and heads. 
This helps us assess how well the {\pretrained} model serves as a starting point for downstream tasks.
Last, we ablate the importance of different components of our {\pretraining} strategy in~\cref{sec:ablations} and verify that downstream performance scales with the amount of unlabeled data in~\cref{sec:scaling}.

\begin{figure}[h!]
    \centering
    \includegraphics[width=\linewidth]{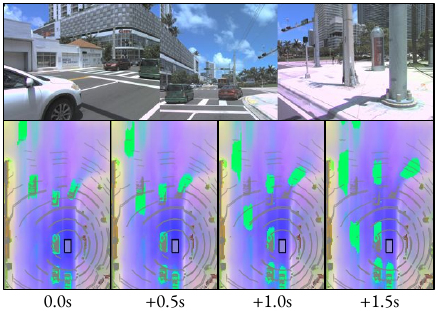}
    \caption{Predicted future VLM features from a Bird's Eye View. The model correctly predicts the car taking a right turn as well as those going straight through the crossing.}
    \label{fig:qual-example-crossing}
\end{figure}

\subsection{Experimental setup and implementation details}
We reimplement UnO~\cite{agro2024uno} as the baseline for our experiments.
To verify the correctness of our implementation, we train and evaluate the model using the training schedule and evaluation protocol reported in the paper, and achieve performance on par with the published results. 
Our applicable improvements, such as better schedule, rotation augmentation, and missing ray supervision, boost performance beyond the originally reported numbers.
For a fair comparison, we use our improved UnO as the baseline in all experiments.%, unless explicitly stated otherwise. 
See~\cref{app:sec:reimplementation} for more details.

We evaluate performance using different amounts of labeled samples $n \in [1, 10^5]$.
For low amounts of labeled data ($n \leq 10^2$), we observe significant variance in performance depending on the samples used during training.
Therefore, we train with 10 different random seeds and report the mean and standard deviation of the evaluation results. At larger sample sizes ($n \geq 10^3$), the variance is negligible.

Unless specified otherwise, the point cloud input range is $x,y \in \pm70$m and $z \in [-2, 6]\, \mathrm{m}$ with a pillar size of $0.16 \times 0.16 \,\mathrm{m^{2}}$.
We use $K_{past}=3$ \lidar~scans at an interval of \SI{0.5}{\second}.
We train for $100,000$ steps with the Adam optimizer~\cite{Kingma2014AdamAM}, a cosine annealing learning rate schedule with a maximum learning rate of $4\cdot10^{-4}$ warming up for $2000$ steps, and an effective batch size of $8$.
We follow~\cite{agro2024uno} and use a buffer size of $\delta=\SI{0.1}{\meter}$ for positive occupancy and DINOv2 queries. % $\delta=0.1$ has only a minor effect on performance~\cite{agro2024uno}
For each training sample, $N_O^+=N_O^-=0.9$M queries are used from $\mathcal{D}_O^+$ and $\mathcal{D}_O^-$ respectively. DINOv2 features are reduced to their $d=16$ principal components, determined on a randomly sampled subset of the training data. The features are cached for each image prior to training. Each sample uses $N_F=100$k queries from $\mathcal{D}_F$. 
%We remove the positional encodings in the raw DINOv2 features are denoised following~\cite{yang2024denoising}.
For ego path we use buffers $w_{ego} = 1$m and sample $N_E^+ = N_E^- = 10$k queries from $\mathcal{D}_E^+$ and $\mathcal{D}_E^-$. 
%We randomly sample $10\%$ of the negative ego path queries from the entire region of interest improve predictions in rarely traversed regions.
Rotation augmentations are sampled from $\theta \in \mathcal{U}(-20\degree, 20\degree)$.
The loss weights are set to $\lambda_{\text{occ}}=1.0$, $\lambda_{\text{dino}}=0.5$, and $\lambda_{\text{ego}}=0.1$.
We train and evaluate our model using the Argoverse 2~\cite{Argoverse2} dataset.
For online mapping, results are based on pre- and post-training on the geographically disjoint splits proposed in~\cite{lilja2024localization} while other tasks use the original training and validation splits.
\begin{figure}[t]
    \centering
    \begin{subfigure}[t]{0.49\linewidth}
        \centering
        \includegraphics[width=\linewidth]{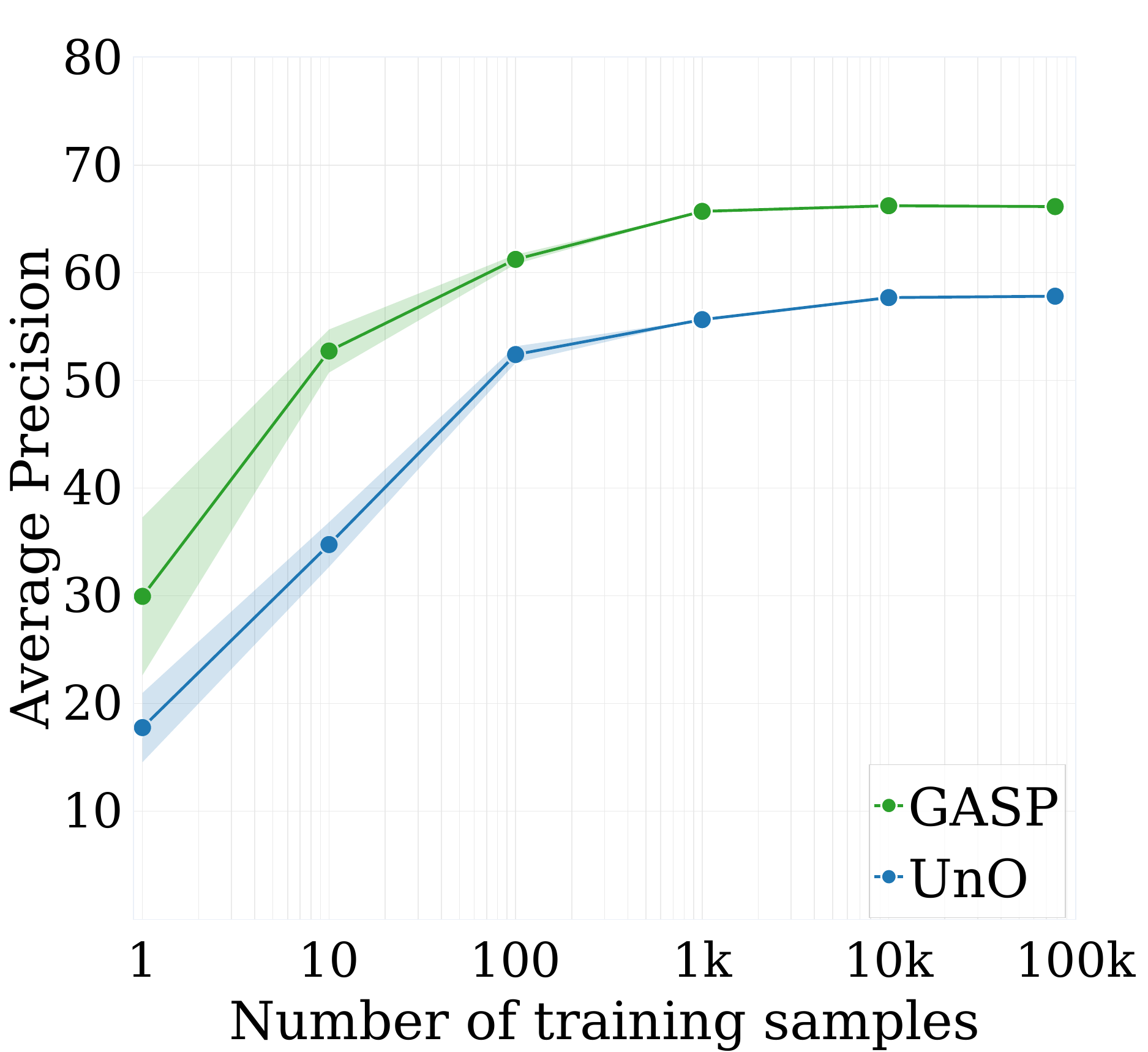}
        \caption{Frozen ({\snow}) encoder.}
        \label{subfig:semfor_frozen}
    \end{subfigure}%
    ~ 
    \begin{subfigure}[t]{0.49\linewidth}
        \centering
        \includegraphics[width=\linewidth]{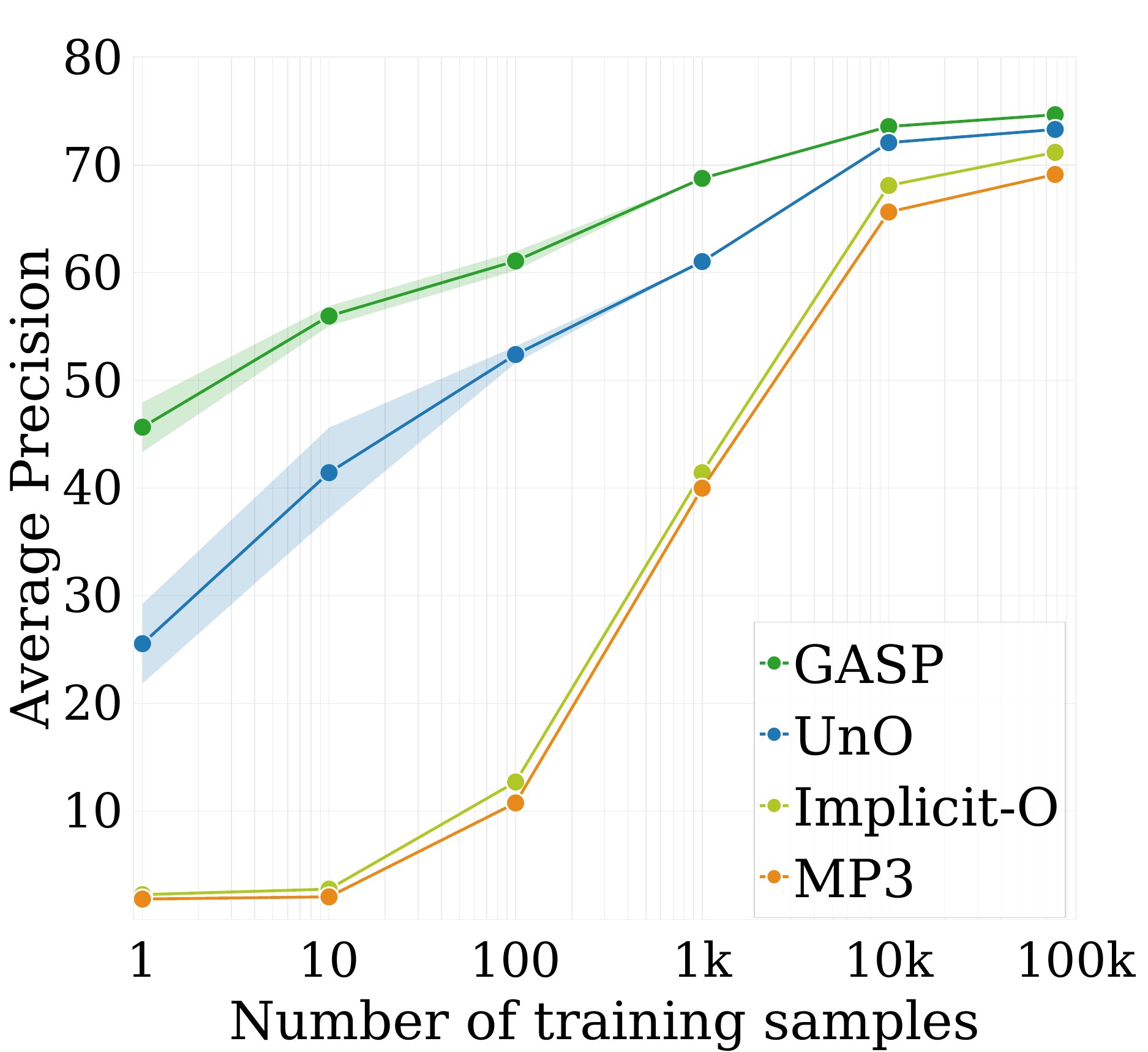}
        \caption{Unfrozen ({\fire}) encoder.}
    \label{subfig:semfor_unfrozen}
    \end{subfigure}
    \caption{Semantic BEV forecasting AP (mean and std. dev) over the number of labeled training samples.} % with the sensor encoder frozen (a) and unfrozen (b).}
    \label{fig:semfor}
    %\vspace{-2mm}
\end{figure}

\subsection{Geometric 4D occupancy forecasting}
\label{sec:experiment-4d-occ}
To evaluate the {\pretrained} model's geometric understanding, we follow~\cite{agro2024uno} and assess its 4D occupancy forecasting performance.
The task is to predict the occupancy of 3D coordinates at future time steps, without any finetuning.

For fair comparison and eliminating the need for manual threshold tuning, we measure recall at a fixed precision of $70\%$.
Predictions are obtained by querying the model over a spatial region of $80 \times 80 \, \mathrm{m}^{2}$ around the ego vehicle, with a uniform sampling interval of \SI{0.2}{\meter} in all spatial directions. 
Temporally, we evaluate at $\{0.6, 1.2, ..., 3.0\} \, \mathrm{s}$ into the future.
Following~\cite{agro2024uno}, we compute precision using \lidar-based ray tracing~\cite{hu2020you} classifying voxels of size $0.2 \, \mathrm{m}^{3}$ as free if traversed by a \lidar~beam before the measured point. Annotated bounding boxes are used to identify points corresponding to objects, labeling them as occupied.

\parsection{Results}
Comparing \modelname~with UnO in~\cref{tab:ablation-loss-terms} shows that the 4D-occupancy recall at precision 70 (R@P70) increases from $79.4\%$ to $81.9\%$. The performance increase primarily stems from the addition of DINOv2 supervision. Ego path supervision seems to slightly decrease the geometric performance.
Intuitively, predicting the future ego path does not require a full understanding of scene geometry. Qualitatively,~\cref{fig:dino-qual} exemplifies the geometric and semantic capabilities of the learned representation at the current timestep, whereas \cref{fig:qual-example-crossing} highlights predictions in to the future.

\begin{figure}[t]
    \centering
    \begin{subfigure}[t]{0.5\linewidth}
        \centering
        \includegraphics[width=\linewidth]{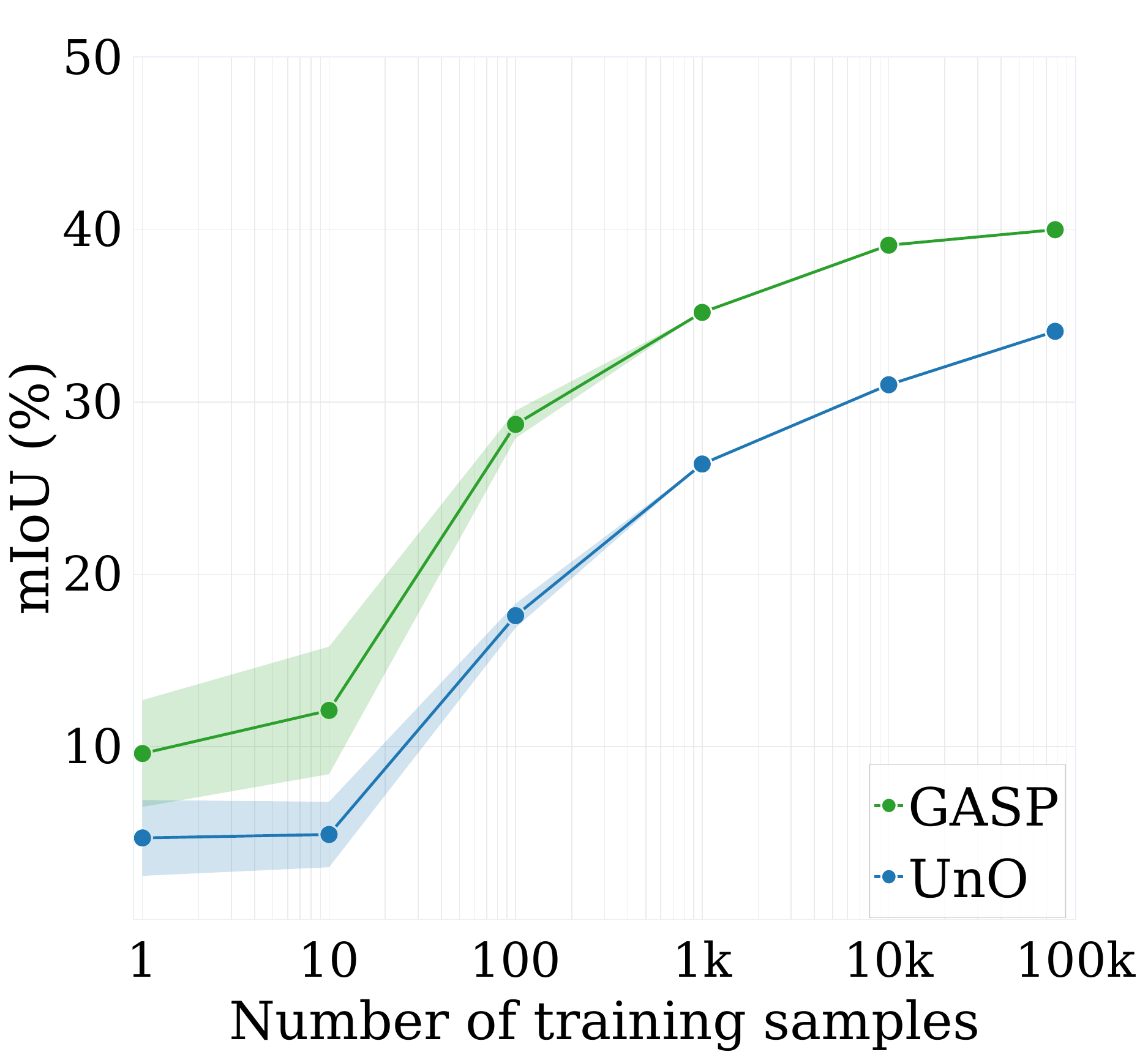}
        \caption{Frozen ({\snow}) encoder.}
        \label{subfig:mapseg_frozen}
    \end{subfigure}%
    ~ 
    \begin{subfigure}[t]{0.5\linewidth}
        \centering
        \includegraphics[width=\linewidth]{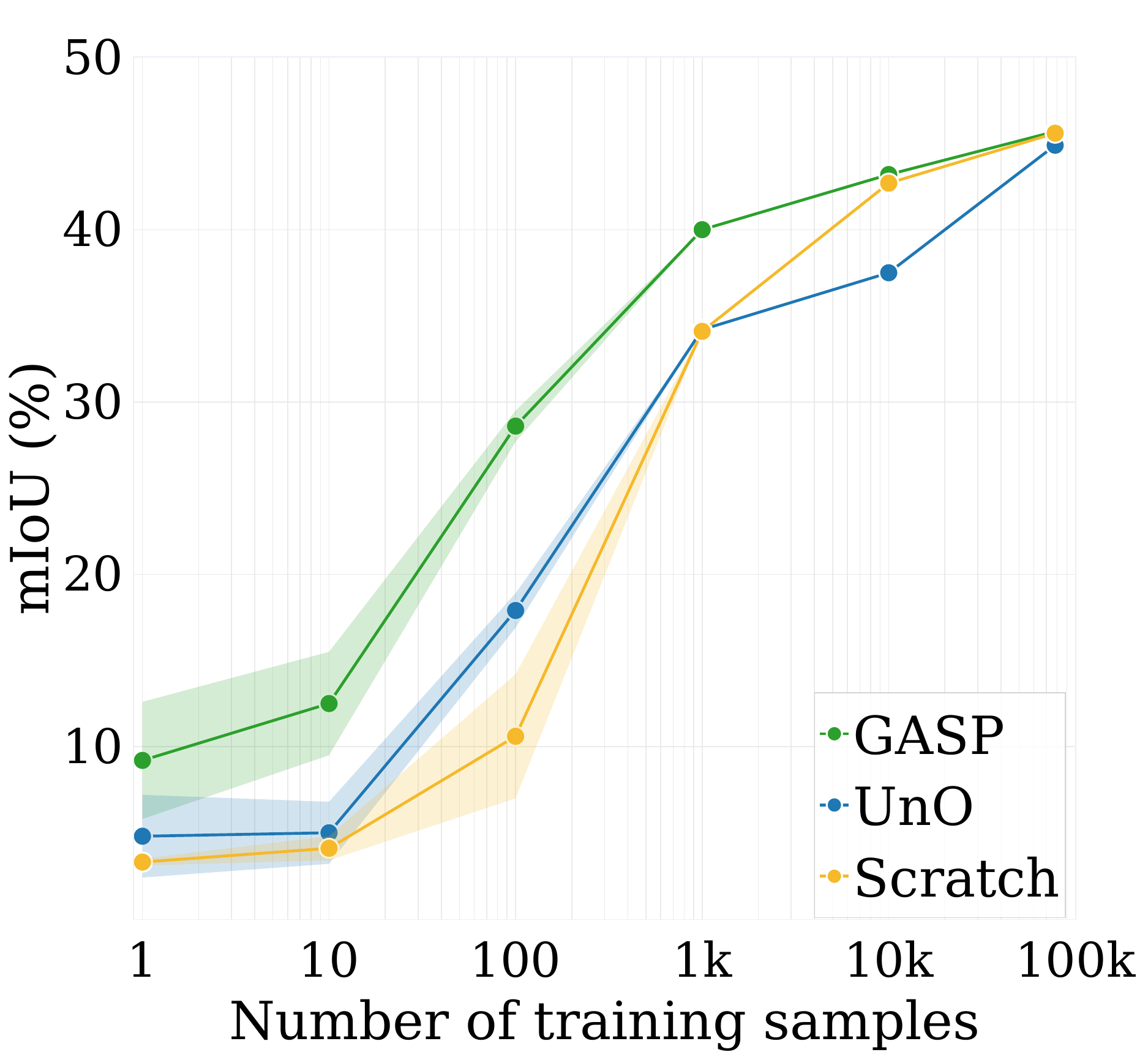}
        \caption{Unfrozen ({\fire}) encoder.}
    \label{subfig:mapseg_unfrozen}
    \end{subfigure}
    \label{fig:mapseg}
    \caption{Map segmentation mIoU (mean and std. dev) across a number of labeled samples with the sensor encoder frozen (a) and unfrozen (b).
    %\modelname~ outperforms the baseline across all amounts of training samples indicating that the BEV features contain a richer BEV representation.
    }
    %\vspace{0mm}
\end{figure}

\subsection{Semantic BEV forecasting}
\label{sec:experiment-semfor}
In Semantic BEV forecasting~\cite{agro2024uno} the model is tasked with forecasting semantic labels and occupancy of 2D coordinates aligned with the ground plane.
We adapt the {\pretrained} occupancy decoder (see~\cref{sec:model-architecture}) to instead predict the occupancy for each class separately.
Following standard protocol~\cite{agro2024uno,agro2023implicit}, we evaluate occupancy for the vehicle class at discrete future times $T = \{\SI{0.0}{\second}, \SI{0.5}{\second},..., \SI{3.0}{\second}\}$ in a uniform grid $80\times80 \, \mathrm{m}^{2}$ centered around the ego-vehicle with a spatial resolution of \SI{0.4}{\meter}.
We measure performance by Average Precision (AP) and Soft-IoU computed across all queries in space and time.

\parsection{Results} We compare our model to the state-of-the-art~\cite{agro2024uno, agro2023implicit,casas2021mp3}. 
In~\cref{fig:semfor}, we show the performance of \modelname~and the UnO baseline for different amounts of labeled samples.
\modelname~consistently outperforms the UnO baseline across all amounts of training samples, demonstrating that the learned representation is more informative for forecasting.
This holds especially true for low amounts of labeled data where \modelname~requires one order of magnitude less data than UnO to reach the same performance.
The gap decreases notably with the amount of labeled samples when the encoder is unfrozen, which is expected given that both models share the same architecture. Performance measured in terms of Soft-IoU follows the same trend, see~\cref{app:semfor}.

\begin{figure}
    \centering
    \includegraphics[width=\linewidth]{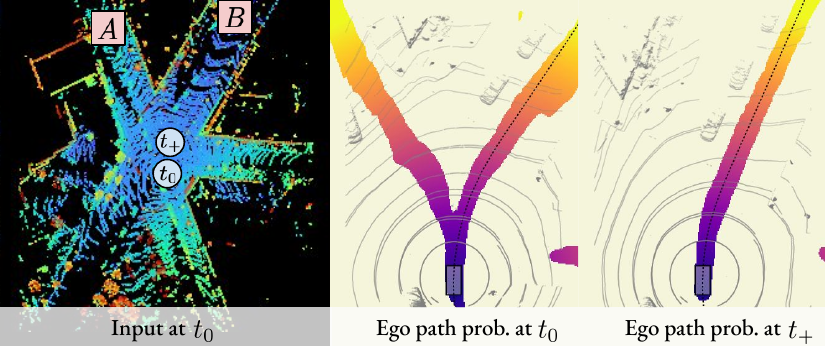}
    \caption{
    Ego path in a crossing, colored by distance to ego-vehicle. 
    \Lidar~point cloud (grey) and true ego path (dashed line) are displayed for reference. 
    At time $t_0$ \modelname~predicts multiple possible modes (A, B), and once it is no longer probable to continue towards A (at $t_+$), the predictions collapse to only one mode. 
    }
    \label{fig:ego-path-dino-crossing}
    \vspace{-3mm}
\end{figure}

\subsection{Map segmentation}
\label{sec:experiment-map-seg}
To assess the semantic content learned from the proposed {\pretraining} scheme, we evaluate its performance on map segmentation.
The task consists of classifying cells in a rasterized grid as lane dividers, road boundaries, or pedestrian crossings, which we predict using a lightweight U-Net-inspired decoder~\cite{harley2023simple} on top of $Z$.
We consider an $80 \times 80 \, \mathrm{m}^{2}$ region around the ego vehicle with a cell size of \SI{30}{\centi\meter}.
We report the mean intersection over union (mIoU) as the evaluation metric.

\parsection{Results}
While \lidar-only map segmentation is underexplored, we note that \modelname~outperforms \SOTA~camera-only setups~\cite{lilja2024localization}.
As shown in~\cref{subfig:mapseg_frozen}, when freezing the encoder and training only the map segmentation head, \modelname~consistently outperforms the baseline across all training set sizes. Our method reaches saturation at $10^4$ training samples, indicating that it learned highly generalizable features. 
The trend persists when unfreezing the encoder, shown in~\cref{subfig:mapseg_unfrozen}.

The results suggest that our {\pretrained} model captures essential features for online mapping, even pedestrian crossings, despite never being trained to detect them. 
The performance gap between the frozen and unfrozen models at $10^5$ training samples is only $5$ mIoU, highlighting that much of the necessary information for map segmentation is encoded during {\pretraining}. For exact metrics, see~\cref{supmat:map-seg}.

\begin{table}[t]
    \centering
    \small
    \setlength{\tabcolsep}{3pt}
    \begin{tabular}{cc  cccccc }
        \toprule
        \multirow{2}{*}{Enc.} & \multirow{2}{*}{{\Pretraining}} & \multicolumn{2}{c}{min$_6$} & \multicolumn{2}{c}{min$_1$}  \\
                              &              & ADE $\downarrow$   & FDE $\downarrow$    & ADE $\downarrow$ & FDE $\downarrow$ \\ \midrule
        \multirow{2}{*}{{\snow}}
                              & UnO          & $0.834$ & $1.43$ & $1.84$ & $3.70$ \\
                              & {\modelname} & $0.617$ & $1.06$ & $1.39$ & $2.87$  \\ \midrule
        \multirow{3}{*}{{\fire}}
                              & -            & $0.880$ & $1.64$ & $1.70$ & $3.58$ \\
                              & UnO          & $0.902$ & $1.51$ & $1.62$ & $3.15$ \\
                              & {\modelname} & $0.706$ & $1.27$ & $1.42$ & $2.74$  \\ 
        \bottomrule
    \end{tabular}
    \caption{Ego-trajectory prediction the full Argoverse 2 sensor dataset, using frozen ({\snow}) and unfrozen ({\fire}) encoders.}
    \label{tab:performance-ego-trajectory-forecasting}
    \vspace{-4mm}
\end{table}
\subsection{Ego-trajectory prediction}
\label{sec:experiment-ego-trajectory}
To evaluate the model’s understanding of the ego vehicle’s future trajectory under our proposed {\pretraining} scheme, we start by inspecting its predicted paths.
In~\cref{fig:ego-path-dino-crossing} the model proposes multiple plausible modes, indicating an awareness of multi-modal future motion and drivable areas.
To further assess its learned motion understanding in a structured geometric representation, including velocity, we employ a simple trajectory decoder as a post-training step. The decoder aggregates information from the feature map, $Z$, via deformable attention~\cite{zhu2020deformable} into latent template trajectories that are later decoded into 2D coordinates with an MLP.
It is trained with an imitation-learning objective between the predicted and recorded trajectory.
%where the goal is to predict valid future trajectories supervised by the driver’s recorded actions.

\parsection{Results}
In~\cref{tab:performance-ego-trajectory-forecasting} we report $\text{minADE}_{1\&6}$, and $\text{minFDE}_{1\&6}$, standard metrics for motion forecasting in Argoverse. These metrics are analogous to L2-planning metrics reported in end-to-end driving works~\cite{hu2023_uniad,weng2024drive}, whilst still allowing for multiple trajectory proposals. The results show that \modelname~captures future ego motion better than UnO in both settings, and significantly outperforms training from scratch, despite using the full amount of trajectory labels.

\begin{table}[t]
    \centering
    \resizebox{0.475\textwidth}{!}{%
    \begin{NiceTabular}{c ccc |c ccc ccc}
    \CodeBefore
      \rowcolor{CornflowerBlue!50}{4}
      \rowcolor{YellowGreen!50}{7}
      \rowcolor{Apricot!50}{8}
      \rowcolor{CornflowerBlue!50}{9}
      \rowcolor{YellowGreen!50}{12}
      % overwrite first col
      \columncolor{white}{1}
    \Body
      \toprule
      \Block{3-1}{Enc.} & \Block{1-3}{\textbf{Components}} & & & \Block{1-1}{\textbf{4D-occ} $\uparrow$} & \Block{1-3}{\textbf{Sem. Forecasting} $\uparrow$} & & & \Block{1-3}{\textbf{Map Seg.} $\uparrow$} & & \\
      & Occ. & E.p. & Sem. & \Block{2-1}{n/a} & \Block{1-3}{Labeled samples} & & & \Block{1-3}{Labeled samples} & & \\
      & & & & & $10^2$ & $10^3$ & $10^5$ & $10^2$ & $10^3$ & $10^5$ \\
      \midrule
      \Block{4-1}{{\snow}} & \checkmark & & & $79.4$                  & $50.3$                                   & $56.7$                                 & $58.4$    & $18.5$ & $27.8$ & $34.1$ \\
      & \checkmark & \checkmark & & $78.9$                  & $60.2$                                   & $62.3$                                 & $63.4$    & $20.9$ & $28.9$ & $35.8$ \\
      & \checkmark & & \checkmark & $81.9$                  & $60.8$                                   & $63.5$                                 & $64.0$    & $22.0$ & $30.2$ & $35.6$ \\
      & \checkmark & \checkmark & \checkmark & $81.6$                  & $59.3$                                   & $64.5$                                 & $64.1$    & $30.6$ & $35.2$ & $40.0$ \\
      \midrule
      \Block{5-1}{{\fire}} & & & & n/a & $19.3$ & $51.7$ & $76.7$ & $10.1$ & $34.1$ & $45.6$ \\
      & \checkmark & & & n/a & $50.5$ & $65.4$ & $76.8$ & $20.2$ & $34.2$ & $44.9$ \\
      & \checkmark & \checkmark & & n/a & $60.1$ & $68.1$ & $77.0$ & $25.4$ & $35.1$ & $43.7$ \\
      & \checkmark & & \checkmark & n/a & $60.6$ & $67.3$ & $77.3$ & $23.7$ & $37.1$ & $43.9$ \\
      & \checkmark & \checkmark & \checkmark & n/a & $59.8$ & $68.3$ & $77.0$ & $29.1$ & $40.0$ & $45.7$ \\
      \bottomrule
    \end{NiceTabular}%
    }
    \caption{Ablation over the {\pretraining} components of \colorbox{YellowGreen!50}{\modelname}. 
    We show performance directly obtained from {\pretraining} (4D occ. P@R70) and its generalization to downstream tasks (Semantic BEV Forecasting and Map Segmentation) when finetuned on different amounts of labeled samples. 
    We ablate each component added to \colorbox{CornflowerBlue!50}{UnO} with the sensor encoder frozen ({\snow}) and unfrozen ({\fire}) as well as performance with \colorbox{Apricot!50}{no \pretraining}.}
    \label{tab:ablation-loss-terms}
    \vspace{-4mm}
\end{table}

\subsection{Ablations}
\label{sec:ablations}
To understand the key contributors to our {\pretraining} strategy, we systematically ablate its components and analyze their individual impact. See~\cref{supmat:addtional-ablations} for details.

\parsection{Loss terms}
We introduce different loss terms incrementally and measure their effect on final performance. 
The results, summarized in~\cref{tab:ablation-loss-terms}, highlight the relative contribution of each term and show a significant boost when combined. 
%We see that using smooth-$L1$ loss does not yield any performance gains~\cref{supmat:additional-ablations-loss}.

\parsection{Rotation augmentation}
We vary the maximum rotation angle for data augmentation to determine its influence on model robustness. The results are presented in~\cref{tab:ablation-rot-aug}. This augmentation yields significant and consistent improvements for both UnO and \modelname~between $\pm5\degree$ and $\pm45\degree$. We opt to use $\pm20\degree$ as a default.
Additionally, we investigate the effect of translation and jitter augmentations, which do not yield meaningful improvements, see~\cref{supmat:addtional-ablations-augmentation}.

\begin{figure}
    \centering
    \includegraphics[width=\linewidth]{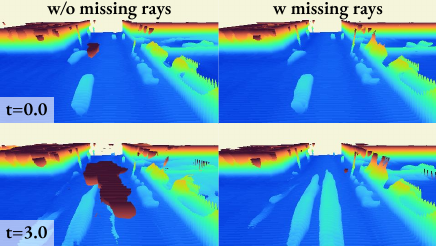}
    \caption{Effect of {\pretraining} with/without missing rays. Without missing rays we observe artifacts in regions where the model is never supervised. Using missing rays as unoccupied supervision, these artifacts are greatly reduced. Geometries are colored by height; blue down and red up.}
    \label{fig:ablation-missing-points}
    \vspace{-4mm}
\end{figure}

\parsection{Missing rays}
We examine the effect of adding supervision for missing \lidar~rays. Quantitative results in~\cref{supmat:additional-ablations-missing-rays}, do not show a significant impact, as the effect of missing rays is not explicitly captured by the current metrics.
However, the primary benefit is visually apparent in \cref{fig:ablation-missing-points}.
Missing ray supervision reduces prediction noise in regions with sparse supervision, such as near region-of-interest boundaries or towards an unobstructed horizon.
Additionally, we note a reduction in occupancy halos above vehicles, which were reported as a failure case in previous work~\cite{agro2024uno}.

\parsection{DINOv2 components}
To evaluate the role of DINOv2 features, we alter the number of components ($8$, $16$, and $32$) and observe the impact on performance. 
We conclude that learning to predict the $16$ most important components yields good results across all three tasks with the full results reported in~\cref{supmat:additional-ablations-dino-dims}.

\begin{table}[t]
    \small
    \centering
    \begin{tabular}{ccccccc}
        \toprule
         \multicolumn{6}{c}{Rotation angle (4D-occupancy $\uparrow$)}                                                                                     \\
                              $\pm0\degree$                              & $\pm5\degree$ & $\pm10\degree$ & $\pm20\degree$ & $\pm45\degree$ & $\pm90\degree$ \\
        \midrule
        $78.1$                                     & $80.1$        & $81.7$         & $81.6$         & $78.1$         & $77.2$         \\
        \bottomrule
    \end{tabular}
    \caption{Rotation augmentation for \modelname. Recall at precision $70\%$ for different angles.}
    \vspace{-2mm}
    \label{tab:ablation-rot-aug}
\end{table}

\subsection{Scaling {\pretraining}}
\label{sec:scaling}
\begin{figure}
    \centering
    \includegraphics[width=\linewidth, trim={0 0 0 17mm}, clip]{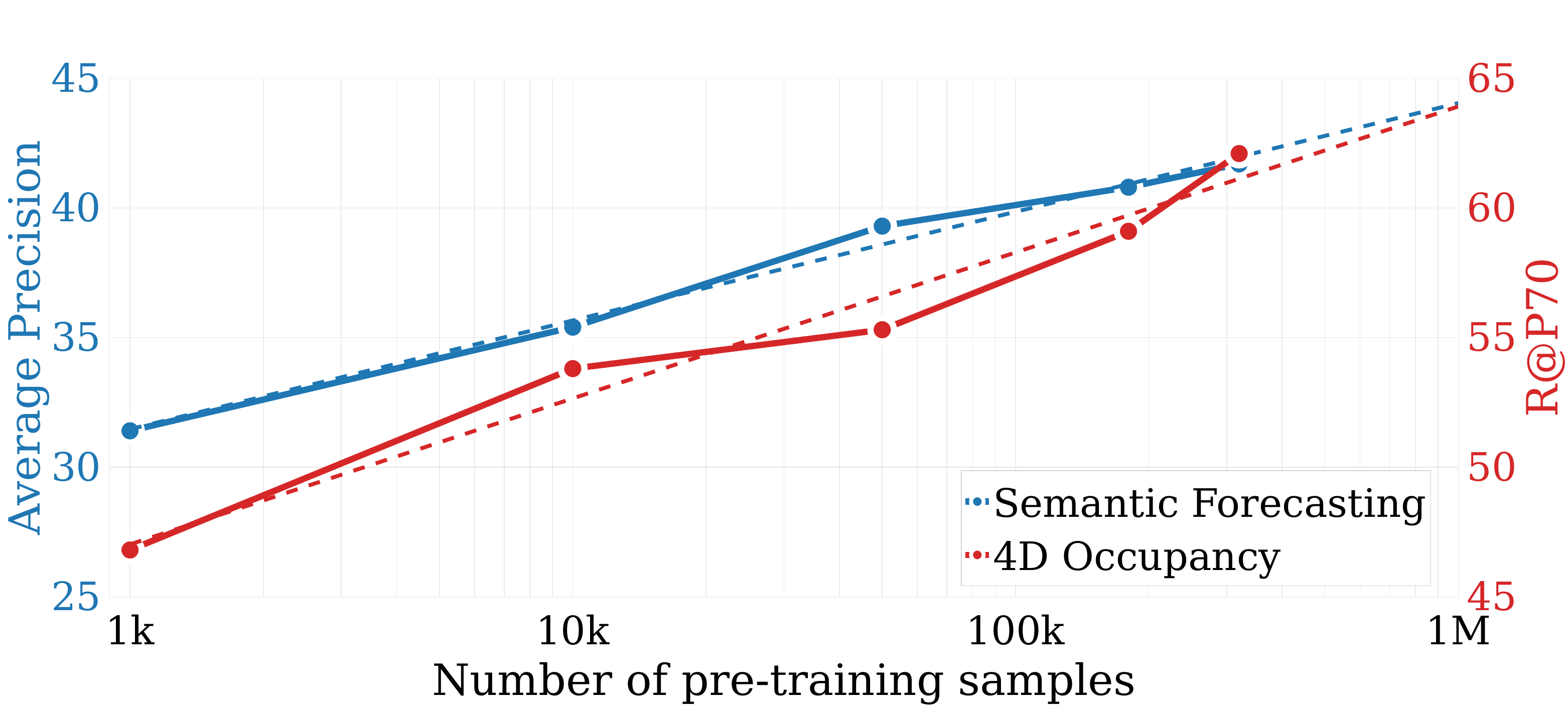}
    \caption{Scaling properties. We vary the number of {\pretraining} samples and evaluate performance (red $\uparrow$) and generalization (blue $\uparrow$), demonstrating {\modelname}'s  remarkably predictable, logarithmic,  scaling behavior.}
    \label{fig:scaling-zod}
    \vspace{-4mm}
\end{figure}

One of the most important qualities of self-supervision is that it continues to show benefits when applied to huge amounts of data. 
To demonstrate this, we train \modelname~on varying number of \textit{\pretraining} samples and evaluate on 4D-occupancy and fine-tuned semantic forecasting (on 1k labeled samples with frozen encoder). Here, we opt to use the Zenseact Open Dataset (ZOD) \cite{alibeigi2023zenseact} to study the scaling behaviour beyond the $\sim$100k training samples available in Argoverse 2 Sensor~\cite{Argoverse2}. The results in \cref{fig:scaling-zod} show that our method scales predictably with no sign of saturation, even when trained on the combined Frames, Sequences, and Drives of ZOD. Furthermore, this experiment shows that \modelname~is dataset-agnostic. The generally lower scores are expected as ZOD has a greater focus on highway driving with higher average velocities than the predominantly inner city driving in AV2.

% \begin{table}[ht]
%     \centering
%     \small
%     \setlength{\tabcolsep}{3pt}
%     \begin{tabular}{cc  cccccc }
%         \toprule
%         \multirow{2}{*}{Enc.} & \multirow{2}{*}{Task} & \multicolumn{5}{c}{{\Pretrained} samples}                                                \\
%         && $1$k& $10$k   & $50$k & $180$k & $320$k   \\ \midrule
%         \multirow{2}{*}{{\snow}}
%                               & 4D Occupancy           & $46.8$ & $53.8$ & $55.3$ & $59.1$  & $62.1$   \\
%                               & Sem. Forecasting & $31.4$ & $35.4$ & $39.3$ & $40.8$ & $41.7$    \\ 
%         \bottomrule
%     \end{tabular}
%     \caption{Scaling properties. We vary the number of {\pretraining} samples and evaluate performance and generalization to show that {\modelname} scales predictably with data.}
%     %\vspace{-2mm}
%     \label{tab:scaling-zod}
% \end{table}

\section{Conclusion}
% introduce
Autonomous driving (AD) generates vast amounts of spatiotemporal data, alluding to the possibility of harnessing scale to learn the underlying geometric and semantic structure of the environment and its evolution over time.
To this end, we introduce \modelname, a self-supervised {\pretraining} strategy that enables scalable representation learning for AD using geometric, semantic, and temporal supervision signals.
%In this work we introduced \modelname~, a geometric, semantic, and temporal self-supervised {\pretraining} strategy that enables scalable representation learning for autonomous driving.
% what we did
Conditioned on past sensory input, \modelname\ is supervised to predict 1) future occupancy, 2) features from a vision foundation model, and 3) ego-path traversal probability at any continuous point $\mathbf{q} = (x, y,z,t)$ in spacetime.
% what does it yield
In doing so, \modelname~learns a rich and generalizable representation of the environment that can be used directly or finetuned for a variety of downstream tasks.
% how does it compare
We demonstrate that our {\pretraining} strategy greatly improves the generalization on tasks such as semantic forecasting, online mapping, and ego-motion forecasting when compared to strategies that only utilize geometric and temporal supervision.
% what are the implications
Our results suggest that \modelname~is a promising approach for learning sensor agnostic and generalizable representations for autonomous driving in a scalable manner. 
We release the code to support further research in this area.

% limitations and future work
%\parsection{Limitations and future work} 
\subsection*{Limitations and future work}
While we only use \modelname\ to \pretrain~a \lidar-based model, the approach is directly applicable to any BEV model, making setups with alternative sensors or complementary multi-modal configurations a promising direction for future work.
Furthermore, leveraging other foundation models (\eg CLIP~\cite{radford2021learning}, SAM~\cite{kirillov2023segany}, or SAL~\cite{ovsep2024better}) or tapping in to other sources of self-supervision (\eg flow-consistency) could further enrich the learned representations. Finally, while \modelname~shows powerful scaling properties, under the current trend we would require roughly 300,000 years of driving to reach near-perfect 4D occupancy prediction -- highlighting the need for further improvements in pre-training efficiency.

%Finally, while \modelname~demonstrates strong scaling properties, the current trend suggest that achieving near-perfect 4D occupancy prediction would require approximately 300,000 years of driving—underscoring the need for more efficient pre-training methods.
\vspace{-1mm}
\subsection*{Acknowledgements}
We thank Georg Hess for fruitful discussions and valuable feedback on the manuscript. We also thank Luca Caltagirone for help with visualizations and Boris Ivanovic insightful discussions and inspiring ideas. This work was partially supported by the Wallenberg AI, Autonomous Systems and Software Program (WASP) funded by the Knut and Alice Wallenberg Foundation. Computational resources were provided by NAISS at \href{https://www.nsc.liu.se/}{NSC Berzelius} and \href{https://www.c3se.chalmers.se/about/Alvis/}{C3SE Alvis}, partially funded by the Swedish Research Council, grant agreement no. 2022-06725.

\newpage
{
    \small
    \bibliographystyle{ieee_fullname}
    \bibliography{main}
}

\newpage
\appendix
    \clearpage
\setcounter{page}{1}
\appendix
\twocolumn[
\centering
\section*{GASP: Unifying Geometric and Semantic Self-Supervised \\ {\Pretrained} for Autonomous Driving}
\subsection*{Appendix}
\vspace{1cm}
]

\section{Baseline reimplementation}
\label{app:sec:reimplementation}
We base our work on the method described in \cite{agro2024uno}. 
However, since their implementation is closed-source we reimplemented their method according to their paper, their predecessor~\cite{agro2023implicit} as well as personal correspondence with the authors~\cite{personalcorresponande}. 
To verify our reimplementation, we report the number of parameters in the original implementation and our version in \cref{tab:app:parameter_count}. 
Note that the decoder head matches perfectly (up to the value number reported in the original paper) and that the encoder is within a $3\%$ margin.
\begin{table}[ht]
    \centering
    \begin{tabular}{l cc}
    \toprule
         \textbf{Parameter count}& Uno encoder & Uno decoder \\ \hline
        Reported & 17.4M & 0.06M \\
        Reimplementation & 16.9M &  0.06M \\
       \bottomrule
    \end{tabular}
    \vspace{-2mm}
    \caption{Parameter count in original UnO (as reported in the paper) and our reimplementation.}
    \label{tab:app:parameter_count}
\end{table}

\begin{figure}[ht]
    \centering
    \includegraphics[width=\linewidth]{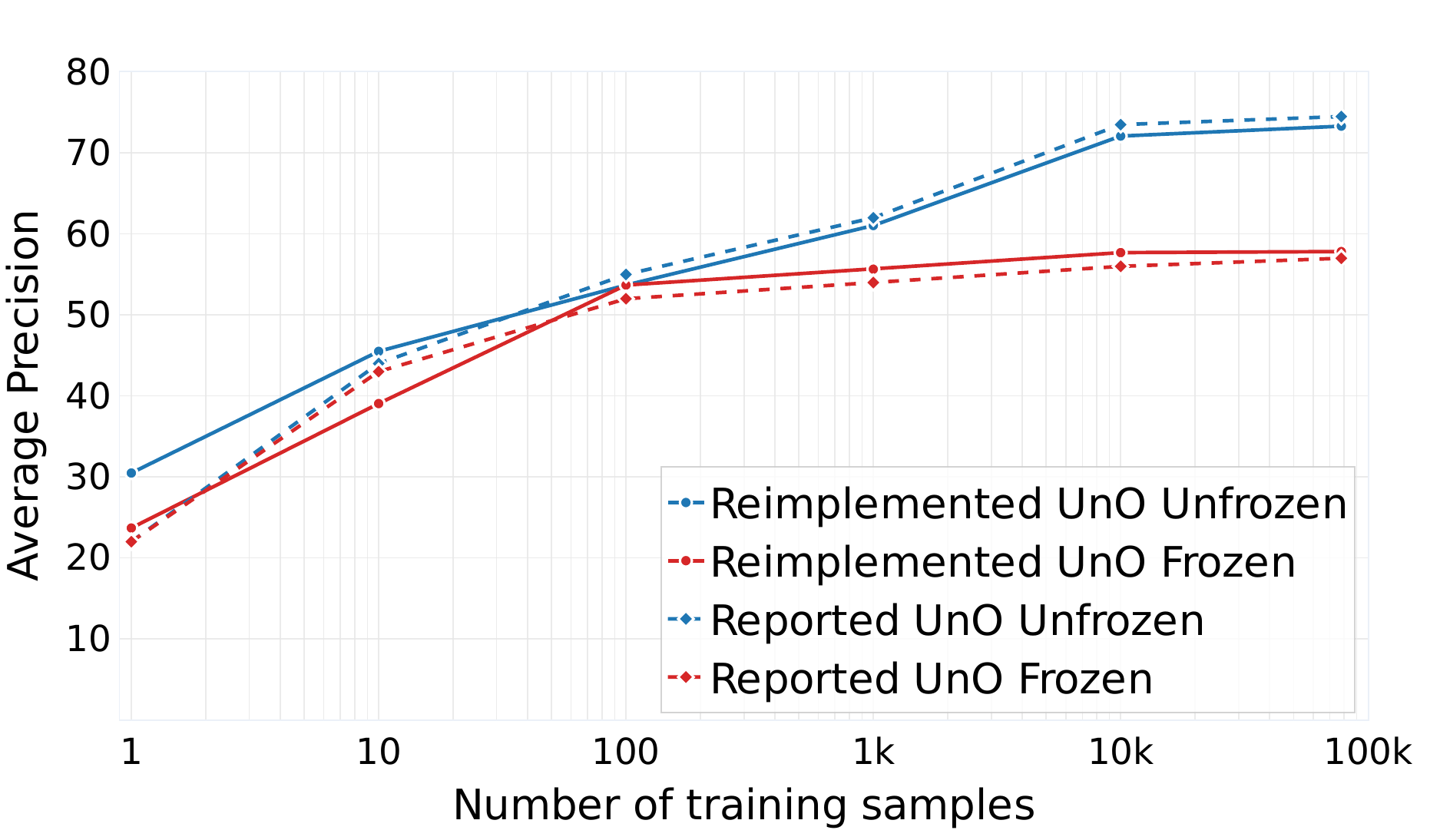}
    \vspace{-6mm}
    \caption{Semantic forecasting AP across number of training samples for our reimplemented baseline (solid) and the results reported in~\cite{agro2024uno} (dashed).}
    \label{fig:app:reimplementation_semfor}
\end{figure}

In addition, we also verify our reimplementation by running identical experiments and comparing to the results reported in the paper. 
In \cref{fig:app:reimplementation_semfor}, we show the \textit{Average Precision} for semantic forecasting across number of training samples for {\finetuning} with both frozen and unfrozen sensor encoder. 

Lastly, we show the results for 4D-occupancy in \cref{tab:reimplement-performance-4d-occ} using recall at precision $70\%$ as the metric, following the original paper~\cite{agro2024uno}. 
Our results show higher performance numbers, which may be due to either improvements in our set-up or slight differences in evaluation settings.
Moreover, applying our training recipe from \cref{sec:experiments} further enhances performance, while additional training improvements, such as rotation augmentation (\cref{sec:rot-aug}) and handling missing rays (\cref{sec:missing-rays}), provide an extra boost. 
For a fair comparison of our main contributions, we refer to the best-performing version as \emph{UnO}.

\begin{table}[ht]
    \centering
    \begin{tabular}{l l cc}
    \toprule
    Name & Details   &R@P $70\%$ \\ \midrule
    UnO$^\dagger$ & reported in \cite{agro2024uno} & $67.0$ \\
    UnO$^r$       & reimplemented                  & $72.0$ \\ \midrule
    UnO$^{+}$     & our training recipe            & $77.6$ \\
    UnO           & our improvements               & $79.4$ \\
    \modelname    & proposed method                & $81.6$ \\
       \bottomrule
        \end{tabular}
    \vspace{-2mm}    
    \caption{4D occupancy recall at precision $70\%$ for our reimplementation and original implementation~\cite{agro2024uno}.}
    \label{tab:reimplement-performance-4d-occ}
\end{table}

\newpage
\section{Additional results}
Here, we show additional results from our evaluation. 
For completeness we also report the numbers visualized as graphs in the main manuscript.
\subsection{Semantic BEV forecasting}
\label{app:semfor}
First, in \cref{tab:performance-semfor-softiou}, we show the performance of {\modelname} and UnO on the semantic BEV forecasting task using the Soft-IoU metric. The results follow the same trend as the Average Precision metrics shown in \cref{tab:performance-semfor-details}.
\begin{table}[ht]
    \centering
    \small
    \resizebox{0.475\textwidth}{!}{
    \setlength{\tabcolsep}{3pt}
    \begin{tabular}{cc  cccccc }
        \toprule
        \multirow{2}{*}{Enc.} & \multirow{2}{*}{PT} & \multicolumn{6}{c}{Labeled data samples}                                                                \\
                              &                            & $1$                                       & $10$            & $10^2$          & $10^3$ & $10^4$ & $10^5$ \\ \midrule
        \multirow{2}{*}{{\snow}}
                              & UnO                        & $11.6^{\pm 1.8}$                          & $14.8^{\pm2.0}$ & $21.3^{\pm1.4}$ & $21.3$ & $22.3$ & $22.6$ \\
                              & {\modelname}               & $19.0^{\pm2.6}$                           & $23.1^{\pm4.5}$ & $28.0^{\pm0.9}$ & $30.5$ & $29.4$ & $31.4$ \\ \midrule
        \multirow{2}{*}{{\fire}}
                              & UnO                        & $16.6^{\pm2.4}$                           & $22.5^{\pm3.9}$ & $25.6^{\pm2.2}$ & $27.4$ & $39.9$ & $39.5$ \\
                              & {\modelname}               & $26.8^{\pm2.2}$                           & $27.9^{\pm1.4}$ & $29.0^{\pm2.8}$ & $36.0$ & $39.6$ & $41.0$ \\
        \bottomrule
    \end{tabular}
    }
    \caption{Semantic BEV forecasting performance (Soft-IoU) for {\modelname} and UnO across different number of {\finetuning} samples with frozen ({\snow}) and unfrozen ({\fire}) sensor encoder.}
    \label{tab:performance-semfor-softiou}
\end{table}

For completeness, we show the detailed numbers from \cref{fig:semfor} in \cref{tab:performance-semfor-details}
\begin{table}[ht]
    \centering
    \small
    \resizebox{0.475\textwidth}{!}{
    \setlength{\tabcolsep}{3pt}
    \begin{tabular}{cc  cccccc }
        \toprule
        \multirow{2}{*}{Enc.} & \multirow{2}{*}{PT} & \multicolumn{6}{c}{Labeled data samples}                                                                \\
                              &                            & $1$                                    & $10$            & $10^2$          & $10^3$ & $10^4$ & $10^5$ \\ \midrule
        \multirow{2}{*}{{\snow}}
                              & UnO                        & $17.8^{\pm 3.2}$                         & $34.8^{\pm 2.1}$ & $52.4^{\pm 0.8}$ & $55.6$ & $57.7$ & $57.8$ \\
                              & {\modelname}               & $30.0^{\pm 7.3}$                         & $52.7^{\pm 2.0}$ & $61.2^{\pm 0.4}$ & $65.7$ & $66.2$ & $66.1$ \\ \midrule
        \multirow{2}{*}{{\fire}}
                              & UnO                        & $25.6^{\pm 3.7}$                        & $41.4^{\pm 4.2}$ & $52.4^{\pm 0.8}$ & $61.1$ & $72.1$ & $73.3$ \\
                              & {\modelname}               & $45.7^{\pm 2.3}$                        & $56.0^{\pm 0.9}$ & $61.1^{\pm 0.9}$ & $68.8$ & $73.6$ & $74.7$ \\
        \bottomrule
    \end{tabular}
    }
    \caption{BEV semantic forecasting performance (AP) showed across number of {\finetuning} samples  with frozen ({\snow}) and unfrozen ({\fire}) sensor encoder.}
    \label{tab:performance-semfor-details}
\end{table}

\subsection{Map segmentation}
\label{supmat:map-seg}
In \cref{tab:performance-mapseg} we report detailed results for the map segmentation task showed in \cref{fig:mapseg}.
\begin{table}[ht]
    \centering
    \small
    \resizebox{0.475\textwidth}{!}{
    \setlength{\tabcolsep}{3pt}
    \begin{tabular}{cc  cccccc }
        \toprule
        \multirow{2}{*}{Enc.} & \multirow{2}{*}{PT} & \multicolumn{6}{c}{Labeled data samples}                                                                \\
                              &                            & $1$                                    & $10$            & $10^2$          & $10^3$ & $10^4$ & $10^5$ \\ \midrule
        \multirow{2}{*}{{\snow}}
                              & UnO                        & $4.7^{\pm 2.2}$                        & $4.9^{\pm 1.9}$ & $17.6^{\pm0.7}$ & $26.4$ & $31.0$ & $34.1$ \\
                              & {\modelname}               & $9.6^{\pm3.1}$                         & $12.1^{\pm3.7}$ & $28.7^{\pm0.8}$ & $35.2$ & $39.1$ & $40.0$ \\ \midrule
        \multirow{3}{*}{{\fire}}
                              & \xmark                     & $3.3^{\pm0.2}$                         & $4.1^{\pm0.7}$  & $10.6^{\pm3.6}$ & $34.1$ & $42.7$ & $45.6$ \\
                              & UnO                        & $4.8^{\pm2.4}$                         & $5.0^{\pm1.8}$  & $17.9^{\pm1.0}$ & $34.2$ & $37.5$ & $44.9$ \\
                              & {\modelname}               & $9.2^{\pm3.4}$                         & $12.5^{\pm3.0}$ & $28.6^{\pm0.9}$ & $40.0$ & $43.2$ & $45.7$ \\

        \bottomrule
    \end{tabular}
    }
    \caption{Map segmentation mIoU (mean and std. dev) across a number of labeled samples with frozen ({\snow}) and unfrozen ({\fire}) sensor encoder. 
    \modelname~ outperforms the baseline across all amounts of training samples indicating that the BEV features contain a richer BEV representation.}
    \label{tab:performance-mapseg}
\end{table}

\subsection{Semantic 4D occupancy}

As an extension to the 3D semantic occupancy (BEV forecasting) in \cref{sec:experiment-semfor} we can post-train the model on also including the height in the prediction, namely 4D (3D+time) occupancy. \cref{tab:performance-sem-4docc} reports the Average Precision performance for UnO and \modelname~ on the vehicle class. We note that \modelname~ outperforms the baseline for all number of labeled samples.\newpage
\begin{table}[ht]
    \centering
    \begin{tabular}{clcccc}
        \toprule
        \multirow{2}{*}{Enc.} & \multirow{2}{*}{PT}& \multicolumn{4}{c}{Labeled data samples} \\
        \cmidrule(lr){3-6}
         & & $100$ & $10^3$ & $10^4$ & $10^5$ \\
        \midrule
        \multirow{3}{*}{{{\fire}}} &Scratch     & $18.4$ & $38.7$ & $44.6$ & $54.7$\\
        &UnO         & $29.1$ & $43.4$ & $45.6$ & $59.9$\\
        &\modelname  & $41.3$ & $47.1$ & $49.0$ & $60.9$\\
        \midrule
    \end{tabular}
    \caption{Average Precision (AP) for Semantic 4D Occupancy Forecasting. We evaluate performance across different numbers of {\finetuning} samples and report results for vehicle segmentation.}
        \vspace{-2mm}
    \label{tab:performance-sem-4docc}
\end{table}

\newpage
\section{Additional ablations}
\label{supmat:addtional-ablations}
For completeness, we present the full results of our ablation studies, highlighting the contribution of each component to the final performance.

\subsection{DINOv2 feature dimensions}
\label{supmat:additional-ablations-dino-dims}
We vary the number of the principal component analysis reduced components ($8$, $16$, and $32$) from DINOv2 that we train to predict.
\cref{tab:ablation-dino-dims} reports the impact on performance. 
We conclude that learning to predict the $16$ most important components yields good results across all three tasks.

\begin{table}[h!]
    \centering
    \resizebox{0.45\textwidth}{!}{%
        \begin{tabular}{cc | c ccc ccc }
            \toprule
            \multirow{3}{*}{Enc.} & \multirow{3}{*}{dim} & \textbf{4D-occ} $\uparrow$& \multicolumn{3}{c}{\textbf{Sem. Forecasting} $\uparrow$} & \multicolumn{3}{c}{\textbf{Map Seg.} $\uparrow$}                                                 \\
                                  &                           &                         & \multicolumn{3}{c}{Labeled samples}   & \multicolumn{3}{c}{Labeled samples}                                     \\
                                  &                           &                         & $10^2$                                   & $10^3$                                 & $10^5$ & $10^2$ & $10^3$ & $10^5$ \\ \midrule
            \multirow{3}{*}{{\snow}}
                                  & $8$                       & $81.1$                  & $50.4$                                   & $54.7$                                 & $54.4$ & $28.5$ & $36.8$ & $40.7$ \\
                                  & $16$                      & $81.6$                  & $59.3$                                   & $64.5$                                 & $64.1$ & $30.6$ & $35.2$ & $40.0$ \\
                                  & $32$                      & $80.2$                  & $59.3$                                   & $62.5$                                 & $63.0$ & $26.9$ & $34.7$ & $38.6$ \\ \midrule
            \multirow{3}{*}{{\fire}}
                                  & $8$                       & n/a                     & $53.4$                                   & $62.7$                                 & $76.4$ & $29.7$ & $39.3$ & $44.7$ \\
                                  & $16$                      & n/a                      & $59.8$                                   & $67.3$                                 & $77.0$ & $29.1$ & $40.0$ & $45.7$ \\
                                  & $32$                      & n/a                      & $60.4$                                   & $67.1$                                 & $76.7$ & $29.1$ & $38.8$ & $45.3$ \\
            \bottomrule
        \end{tabular}
    }
    \caption{Performance across different downstream task when varying the number of DINOv2 dimensions used in the regression objective.}
    \vspace{-5mm}
    
    \label{tab:ablation-dino-dims}
\end{table}

\subsection{Missing rays}
\label{supmat:additional-ablations-missing-rays}
Here, we ablate the use of inferred missing rays during {\pretraining}. 
We measure performance on geometric 4D occupancy using the recall at precision $70\%$ and report the numbers in \cref{tab:performance-missing-rays}. 
As noted in the main manuscript, we do not see any quantitative improvements, but rather qualitative ones. 
We hypothesize that this is because the metric inherently disregards the regions where this supervision helps.

\begin{table}[ht]
    \centering
\begin{tabular}{l c c}
\toprule
Missing rays & \xmark & \cmark \\ \midrule
\modelname & $81.6$ & $81.0$ \\
\bottomrule
\end{tabular}
    \caption{Missing rays. Recall at precision $70\%$}
    \label{tab:performance-missing-rays}
    \vspace{-5mm}
\end{table}

\subsection{Augmentation}
\label{supmat:addtional-ablations-augmentation}
In addition to the rotation augmentation outlined in the main manuscript, we also experiment with translation, jitter augmentations, and the number of feature dimensions in the DINOv2 features. 
We measure geometric 4D occupancy performance as measured by the recall at precision $70\%$.

\parsection{Rotation augmentation}
For completeness, we, apart from \modelname, also show that UnO benefits from rotation augmentation in \cref{tab:supmat-ablation-rot-aug}.

\begin{table}[t]
    \small
    \centering
    \begin{tabular}{l ccccccc}
        \toprule
        \multirow{2}{*}{PT } & \multicolumn{6}{c}{Rotation augmentation (4D-occupancy $\uparrow$)}                                                                                     \\
                             & $\pm0\degree$                              & $\pm5\degree$ & $\pm10\degree$ & $\pm20\degree$ & $\pm45\degree$ & $\pm90\degree$ \\
        \midrule
        UnO                  & $77.6$                                     & $79.2$        & $78.3$         & $79.4$         & $78.7$         & $74.1$         \\
        \modelname           & $78.1$                                     & $80.1$        & $81.7$         & $81.6$         & $78.1$         & $77.2$         \\
        \bottomrule
    \end{tabular}
    \caption{Rotation augmentation. Recall at precision $70\%$}
    \vspace{-2mm}
    
    \label{tab:supmat-ablation-rot-aug}
\end{table}

\parsection{Translation augmentation}  
Similarly to the rotation augmentation, we can augment the training data such that the vehicle is translated in $x$ and $y$  directions.
We ablate the effects of adding such translation augmentation, and \cref{tab:performance-trans-aug} displayed the recall at precision $70\%$. 
We don't see any major improvements using this augmentation and opt to use the more impactful rotation augmentation.
We hypothesize that the data already includes a large variety of lateral and longitudinal shifts.

\begin{table}[ht]
    \centering
\begin{tabular}{l cccc}
\toprule
\multirow{2}{*}{PT } & \multicolumn{4}{c}{Translation augmentation } \\
& $-$ & $\pm 0.5$m & $\pm1.5$m & $\pm3.0$m  \\
\midrule
\modelname & $78.1$ & $79.0$ & $78.3$ & $76.0$  \\
\bottomrule
\end{tabular}
    \caption{Translation augmentation. Recall at precision $70\%$}
        \vspace{-2mm}
    \label{tab:performance-trans-aug}
\end{table}

\parsection{Jitter augmentation} We also experiment with jitter augmentation, which aims to up-sample negative queries close to the positive queries by adding a jitter parameter $\tau$ to the negative query equation:
\begin{equation}
    \mathbf{q}_i^{-} = \mathbf{o}_i + (\mathbf{p}_i - \mathbf{o}_i)d^\tau
\end{equation}
where $d \sim \mathcal{U}(0, 1)$.

Our initial intuition—that increasing jitter would lead to sharper geometry learning—did not hold, as performance declines with higher jitter values. Conversely, reducing jitter also appears to negatively impact model performance.

\begin{table}[ht]
    \centering
\begin{tabular}{l cccccc}
\toprule
\multirow{2}{*}{PT } & \multicolumn{6}{c}{Jitter } \\
& $0.6$ & $0.8$ & $1.0$ & $1.2$ & $2.0$ & $3.0$  \\
\midrule
\modelname & $80.0$ & $80.1$ & $81.6$ & $80.7$ & $75.0$ & $68.5$ \\
\bottomrule
\end{tabular}
    \caption{Jitter. Recall at precision $70\%$}
        \vspace{-2mm}
    
    \label{tab:performance-rot-aug}
\end{table}

\subsection{DINO loss}
\label{supmat:additional-ablations-loss}
We ablate the loss function used to learn our DINOv2 features in \cref{tab:performance-dino-soft-l1}. We again compare performance on geometric 4D occupancy using the recall at precision $70\%$ metric. The Smooth-L1 loss reduces the performance and we opt to use the L1-loss.
\begin{table}[ht]
    \centering
\begin{tabular}{l cccc}
\toprule
\multirow{2}{*}{PT } & \multirow{2}{*}{L1} & \multicolumn{3}{c}{Smooth-L1 ($\beta$)} \\
&  & $0.1$ & $0.5$ & $1.0$   \\
\midrule
\modelname & $81.6$ & $78.9$ & $79.3$ & $79.3$  \\
\bottomrule
\end{tabular}
    \caption{Dino-loss. Smooth L1. Recall at precision $70\%$}
        \vspace{-2mm}
    
    \label{tab:performance-dino-soft-l1}
\end{table}

\newpage
\if False
\section{Other sensor modalities}
\label{app:multisensor}
We show that our method is general enough to handle other sensor setups. Here we show the performance on downstream tasks for different sensor setups. For the camera-only implementation ($C$) we use SimpleBEV~\cite{harley2023simple} to encode surround images to a BEV representation. For the fusion implementation ($L+C$) we simply combine the BEV features obtained from the camera-branch (SimpleBEV) and the LiDAR branch (voxelization + ResNet) with summantion. We note that the performance degrades in the fusion setting (LiDAR + camera), and accredit this to the naïve fusion approach. Future work should investigate the fusion setting for a more sophisticated pipeline (\eg BevFusion~\cite{liu2023bevfusion}), but this is out of scope for this work.
\begin{table}[ht]
    \centering
    \begin{tabular}{ccc ccc}
    \toprule
       Encoder  & Sensors & Semfor & Mapseg & 4D-occ\\ \midrule
       \multirow{3}{*}{{\snow}} 
        & L   & $62.38$ & $28.7$ & $81.6$  \\
        & C   & $42.70$ & $14.0$ & $7.1$   \\
        & L+C & $60.89$ & -      & $77.8$  \\ 
       % \midrule
       % \multirow{3}{*}{{\fire}}   
       %  & L   & $63.08$ & $28.6$ & -  \\
       %  & C   & $44.49$ & - & -   \\
       %  & L+C & $60.99$ & - & -  \\
       \bottomrule
    \end{tabular}
    \caption{Performance using 100 samples for {\finetuning}.}
    \vspace{-2mm}
    
    \label{tab:performance-enc}
\end{table}
\fi

% \section{Missing rays inference details}
% \label{app:missing-rays}

\section{Additional visualizations}
\begin{figure*}
    \centering
    \includegraphics[width=\linewidth]{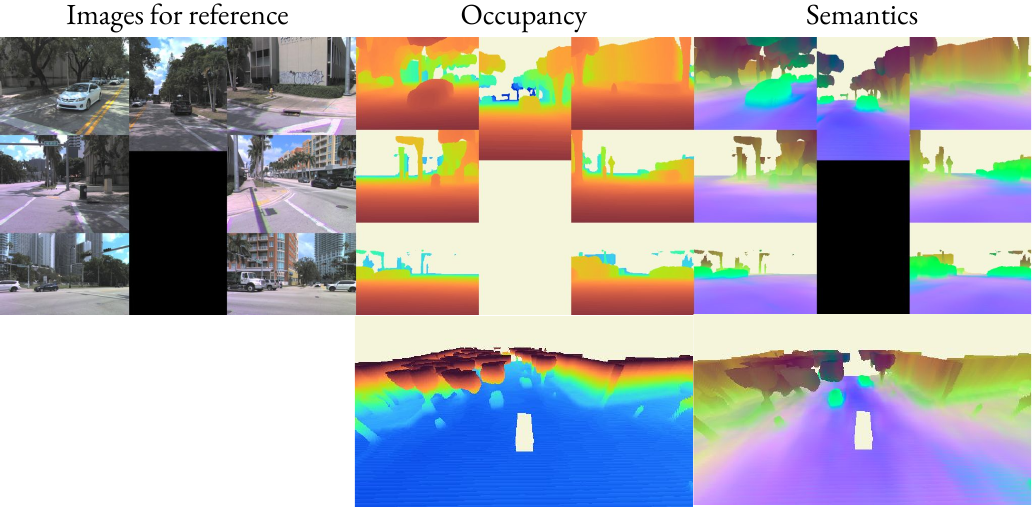}
    \caption{Occupancy and Dino features projected into camera view. Note that the white-box representing ego vehicle has been injected for illustrative purposes.}
    \label{fig:dino-qual-v2}
\end{figure*}

\begin{figure*}
    \centering
    \includegraphics[width=\linewidth]{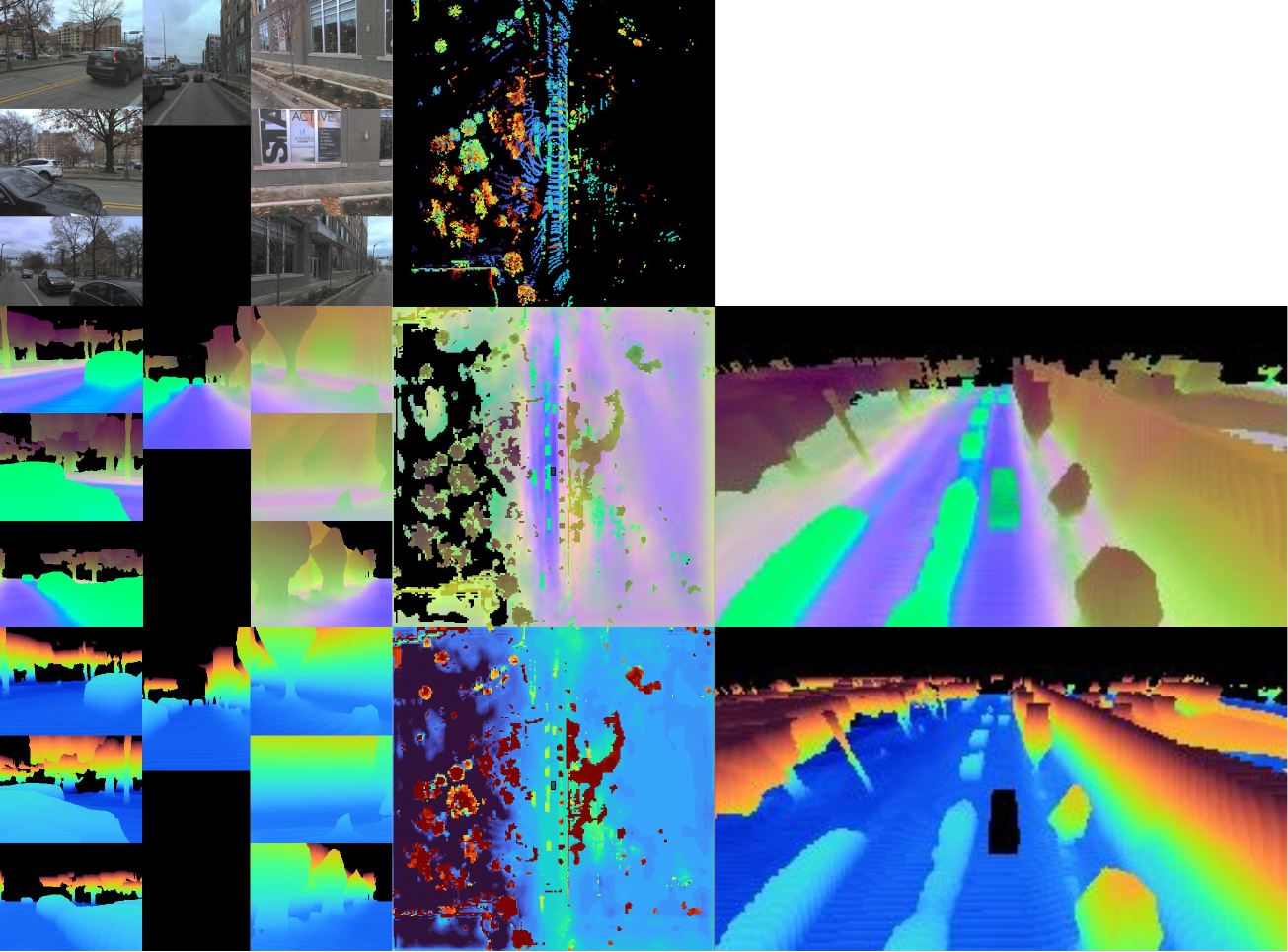}
    \caption{Occupancy and Dino features projected into a camera view, a holistic view, and a bird's-eye view.}
    \label{fig:dino-occupancy-sup}
\end{figure*}

\begin{figure*}
    \centering
    \includegraphics[width=\linewidth]{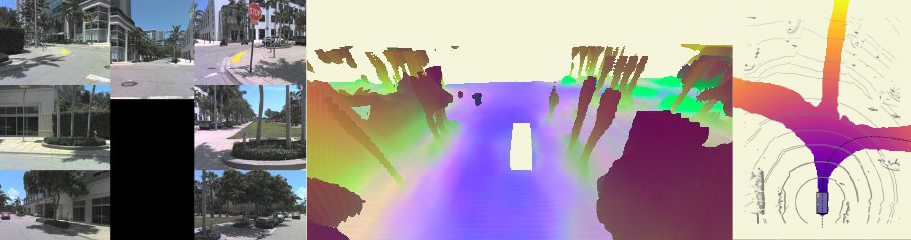}
    \caption{Dino features and ego path in a three-way intersection.}
    \label{fig:qualitative_example_img_dino_path}
\end{figure*}

\begin{figure*}
    \centering
    \includegraphics[width=\linewidth]{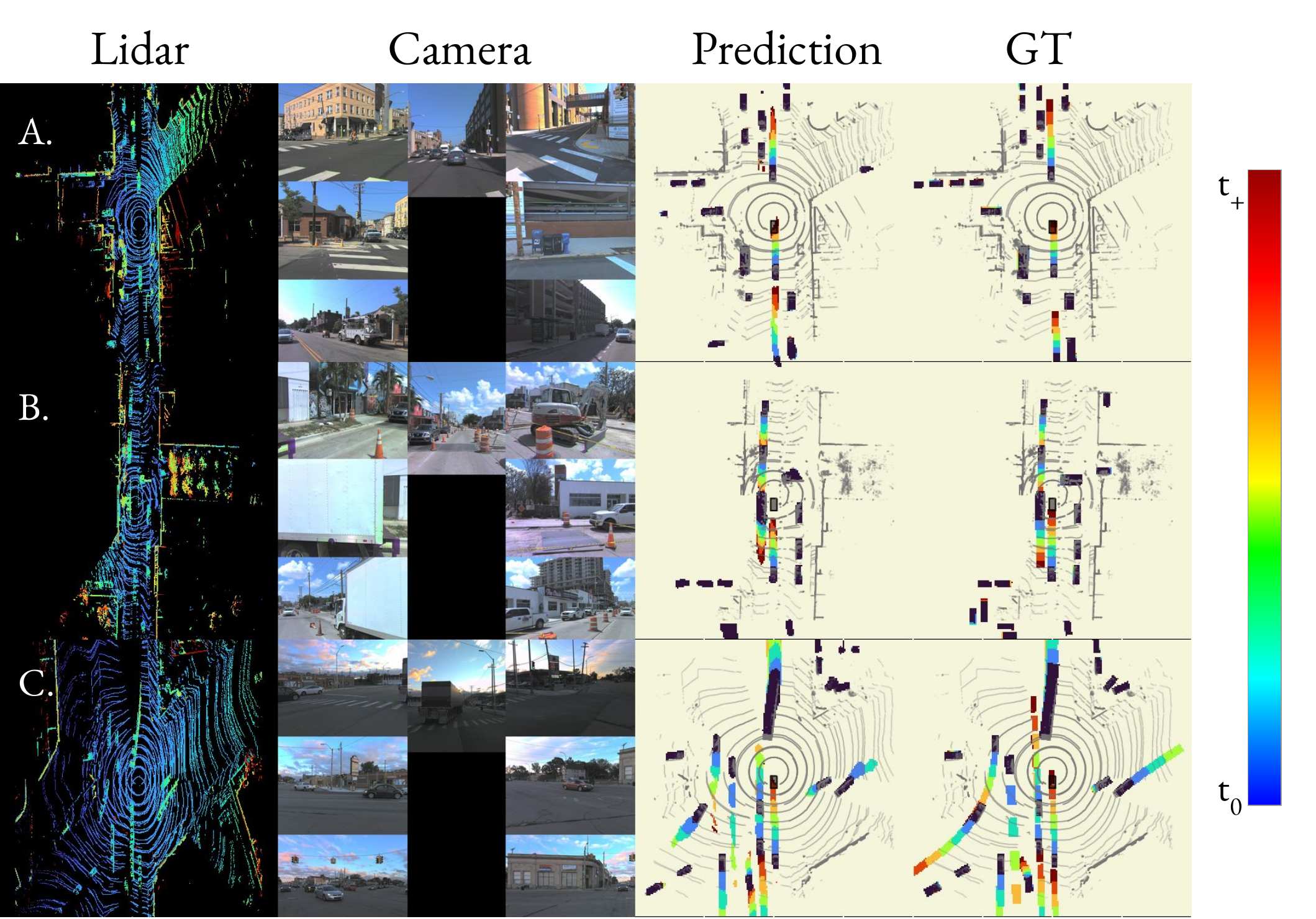}
    \caption{BEV segmentation forecasting results showing A) a typical scenario, B) a scenario with uncommon road users (in this case an excavator), and C) a more complex scenario. An unfrozen \modelname~representation with 100 samples available in the post-training task is used.}
    \label{fig:semfor-sup}
\end{figure*}

\begin{figure*}
    \centering
    \includegraphics[width=\linewidth]{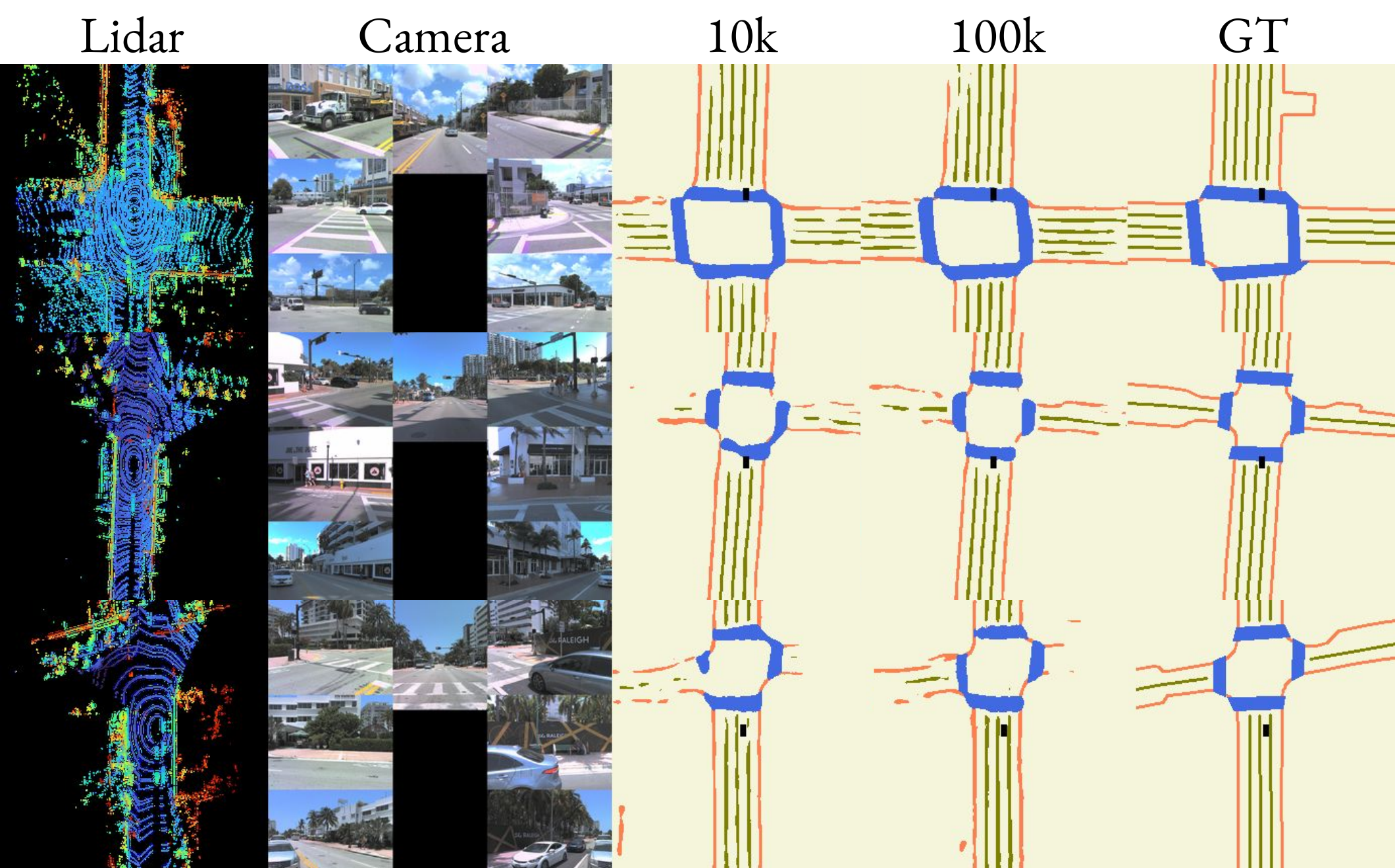}
    \caption{Prediction results from the map segmentation post-training task of \modelname~with frozen encoder. Note that predictions are only made based on lidar input. Camera images are only provided as visual clarity for the reader regarding what scene is being predicted.}
    \label{fig:map_seg_sup}
\end{figure*}

\begin{figure*}
    \centering
    \includegraphics[width=\linewidth]{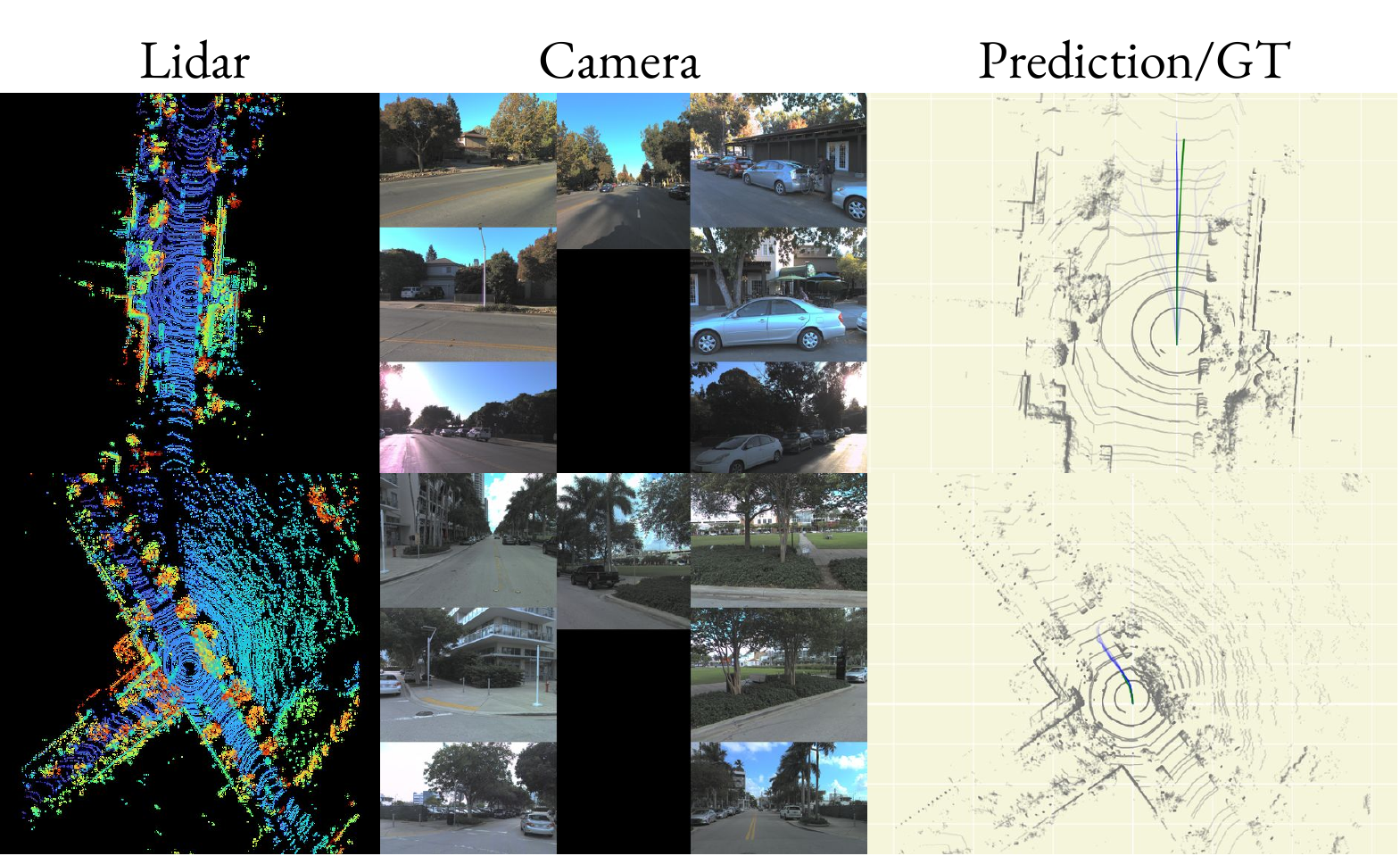}
    \caption{Ego-trajectory prediction results using a frozen \modelname~representation. Expert trajectories, the groundtruth, are shown in green while predictions are shown in blue. Note that camera inputs are only provided as visual support for the reader and are not part of the prediction.
    }
    \label{fig:traj_pred_sup}
\end{figure*}

\begin{figure*}
    \centering
    \includegraphics[width=\linewidth]{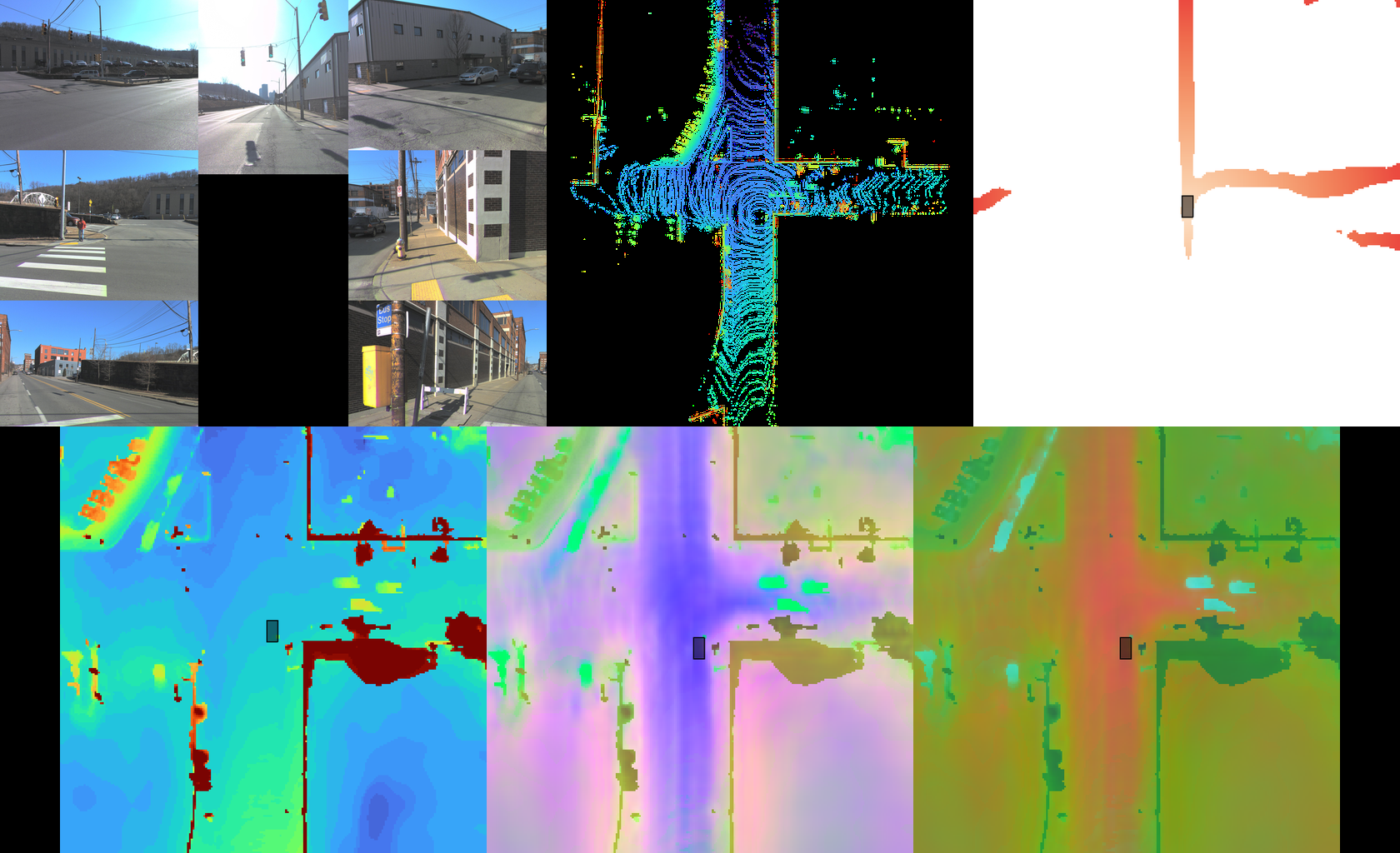}
    \caption{Ego path along with the first and second three most important features, highlighting the complementing aiding properties of the ego path task and the DINO feature prediction task in encoding information about drivable area in the representation.}
    \label{fig:dino-path-qual-sup}
\end{figure*}

\begin{figure*}
    \centering
    \includegraphics[width=\linewidth]{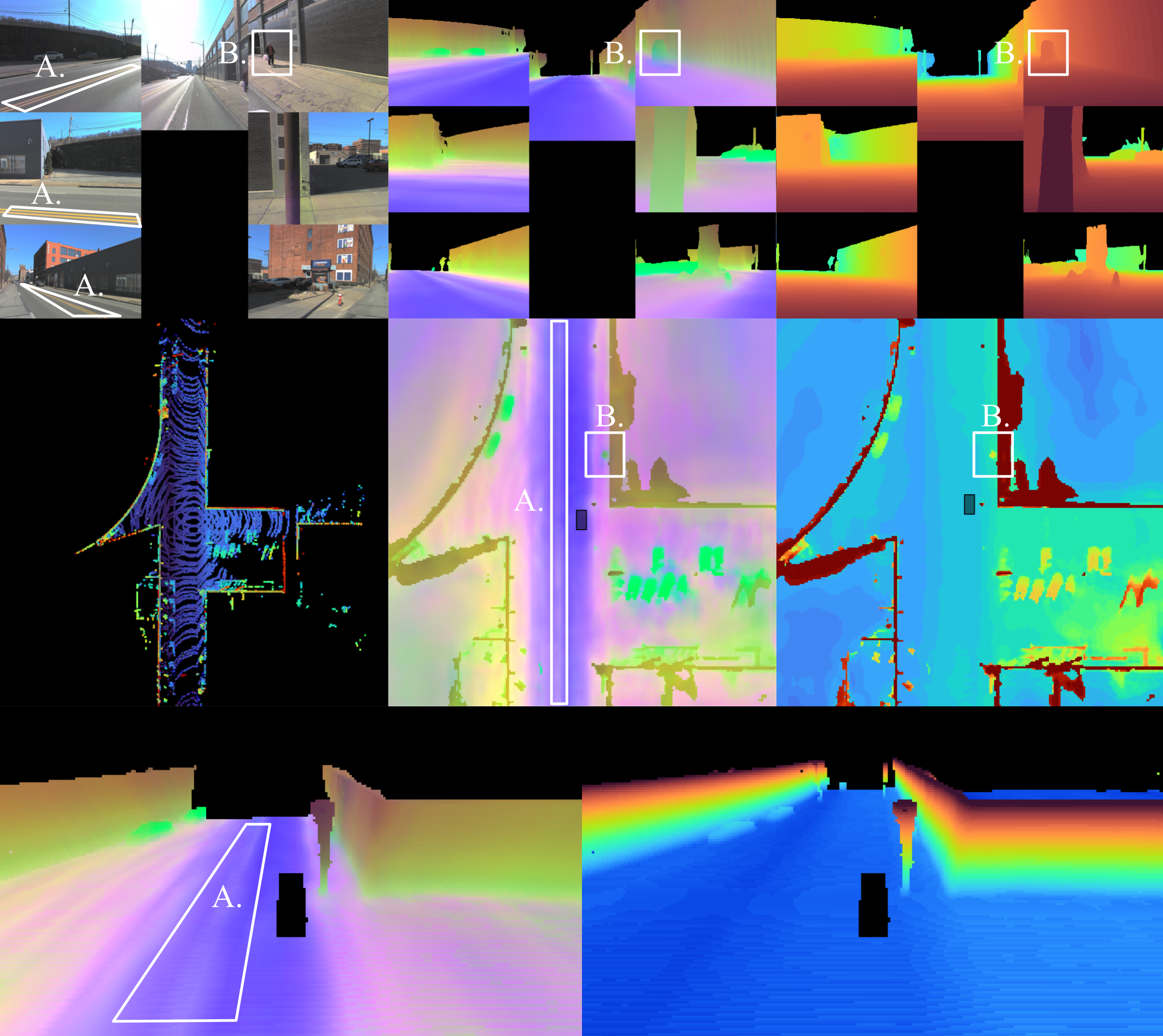}
    \caption{A qualitative example of the feature-level information predicted by the representation produced by \modelname, showcasing its capability of contrasting otherwise diffuse scene elements such as the lane-dividers (marked A.) or the person carrying a bag (marked B.) from the background.}
    \label{fig:dino-occupancy-lane-markings}
\end{figure*}

We provide some additional qualitative results in \cref{fig:dino-qual-v2,fig:dino-occupancy-sup,fig:dino-path-qual-sup,fig:qualitative_example_img_dino_path,fig:map_seg_sup,fig:semfor-sup,fig:traj_pred_sup,fig:dino-occupancy-lane-markings}. In short, they aim to give more examples of the quality of the information that the representation embeds, but also to depict some interesting emergent properties of our {\pretraining} strategy. In \cref{fig:dino-qual-v2} one can view the full holistic view of both semantic and occupancy-field information, as opposed to in \cref{fig:dino-qual}, and in \cref{fig:dino-occupancy-sup} a bird's-eye view of point-cloud inputs, features, and occupancy, is provided. Furthermore, in \cref{fig:qualitative_example_img_dino_path}, the multimodal outputs on path probabilities in a three-way intersection are depicted.

For the BEV semantic forecasting task, \cref{sec:experiment-semfor}, complementary qualitative results are provided in \cref{fig:semfor-sup} for three scenes, a common case, a case with unusual objects, and a more complex case. In \cref{fig:map_seg_sup}, outputs from the map segmentation task are given for a different number of samples in the training set given a frozen \modelname~encoder. Here, we may visually make note of the models ability to predict pedestrian crossings, as indicated in \cref{sec:experiment-map-seg}, despite the absence of this information in the Lidar input data. Qualitative results connected to the ego-trajectory prediction task can be viewed in \cref{fig:traj_pred_sup}.

In \cref{fig:dino-path-qual-sup} we note the similarity between the predicted semantic regions of, what could possibly be understood as, drivable area with the predicted ego future path. Potentially, their joint supervision signal amplifies tasks directly dependent on this type of information, such as map segmentation, which could explain why having both, and not one or the other, seems beneficial for said task. Finally, in \cref{fig:dino-occupancy-lane-markings}, we show that feature prediction in \modelname~has the potential of capturing fine-grained, yet important, scene details to an extent that its geometric occupancy head, and by extension a model which only models geometric occupancy, hardly highlights.  

% \begin{figure*}
%     \centering
%     \includegraphics[width=\linewidth]{assets/qual_examples/ego path.pdf}
%     \caption{Predicted ego path is multi-modal until we have decided where to go...}
%     \label{fig:enter-label}
% \end{figure*}
\end{document}